\documentclass[journal]{IEEEtran}
\usepackage{cite}
\usepackage[pdftex]{graphicx}
\usepackage{subfigure}
\usepackage{algorithm}
\usepackage{algorithmic}
\usepackage{amsmath,amssymb,amsopn,amstext,amsfonts}
\DeclareGraphicsExtensions{.pdf,.eps,.png}
\usepackage{multirow}
\usepackage{array}
\usepackage{url}
\usepackage{color}
\usepackage{mathtools}
\usepackage{booktabs}
\usepackage{multirow}
\usepackage{bm}
\usepackage{diagbox}
\usepackage{supertabular}

\usepackage{balance}
\usepackage[outdir=./eps_pdf/]{epstopdf}
\usepackage[justification=centering]{caption}  

\graphicspath{{./figures_pdf/},...}

\begin{document}
	\title{Control of a Tail-Sitter VTOL UAV \\ Based on Recurrent Neural Networks
	}

	\author{Jinni Zhou\textsuperscript{1},
		Hao Xu\textsuperscript{1},
		Zexiang Li\textsuperscript{1},
		Shaojie Shen\textsuperscript{1},
		Fu Zhang\textsuperscript{2}
		
		\thanks{\textsuperscript{1}Department of Electronic and Computer Engineering, Hong Kong University of Science and Technology (HKUST). email: \{eejinni, hao.xu, eezxli, eeshaojie\}@ust.hk.}
		
		\thanks{\textsuperscript{2}Department of Mechanical Engineering, University of Hong Kong. email:fuzhang@hku.hk.}}
	
	\maketitle

	\begin{abstract}
		Tail-sitter vertical takeoff and landing (VTOL) unmanned aerial vehicles (UAVs) have the capability of hovering and performing efficient level flight with compact mechanical structures. We present a unified controller design for such UAVs, based on recurrent neural networks. An advantage of this design method is that the various flight modes (i.e., hovering, transition and level flight) of a VTOL UAV are controlled in a unified manner, as opposed to treating them separately and in the runtime switching one from another. The proposed controller consists of an outer-loop position controller and an inner-loop attitude controller. The inner-loop controller is composed of a proportional attitude controller and a loop-shaping linear angular rate controller. For the outer-loop controller, we propose a nonlinear solver to compute the desired attitude and thrust, based on the UAV dynamics and an aerodynamic model, in addition to a cascaded PID controller for the position and velocity tracking. We employ a recurrent neural network (RNN) to approximate the behavior of the nonlinear solver, which suffers from high computational complexity. The proposed RNN has negligible approximation errors, and can be implemented in real-time (e.g., 50 Hz). Moreover, the RNN generates much smoother outputs than the nonlinear solver. We provide an analysis of the stability and robustness of the overall closed-loop system. Simulation and experiments are also presented to demonstrate the effectiveness of the proposed method.         
	\end{abstract}
	
	
	\section{Introduction}
	Nowadays, new designs and configurations of small-scale unmanned aerial vehicles (UAVs)  are being developed to meet the higher requirements for professional fields such as parcel delivery, remote sensing, mapping, surveillance and dangerous missions. These applications need the UAVs to have the properties of both power efficiency and maneuverability. In many cases, suitable sites for taking off and landing are limited, so the UAV must have the ability to take off and land vertically. A simple way to design this kind of aircraft is to combine a quad-rotor and fixed-wing UAV called a vertical takeoff and landing (VTOL) UAV.  Among the mainstream VTOL UAVs\cite{zhang2017modeling}, a tail-sitter VTOL UAV is the simplest because it does not need any additional actuators. Simple mechanisms are useful for small-scale UAVs because they can save manufacturing complexity and weight. Therefore, a large number of researchers have been attracted to study tail-sitter VTOL UAVs: in \cite{stone2001t}, a twin-engine design is presented; in  \cite{chu2009automatic}, a structure with coaxial contra-rotating propellers is described; and in \cite{oosedo2013development}, a quad-rotor configuration is developed, to name just a few. Generally, a tail-sitter VTOL UAV has two flight modes: hovering and level flight, and it switches between these two modes by tilting the pitch angle of the aircraft by 90 degrees. To avoid apparent altitude drop or gain, the control of transition flight should be stable and reliable. However, transition control is the most difficult part as it is usually aerodynamically unstable due to the well-known wing stalling phenomenon at a large angle of attack.
	
	\begin{figure}
		\centering
		\includegraphics[width=0.3\textwidth]{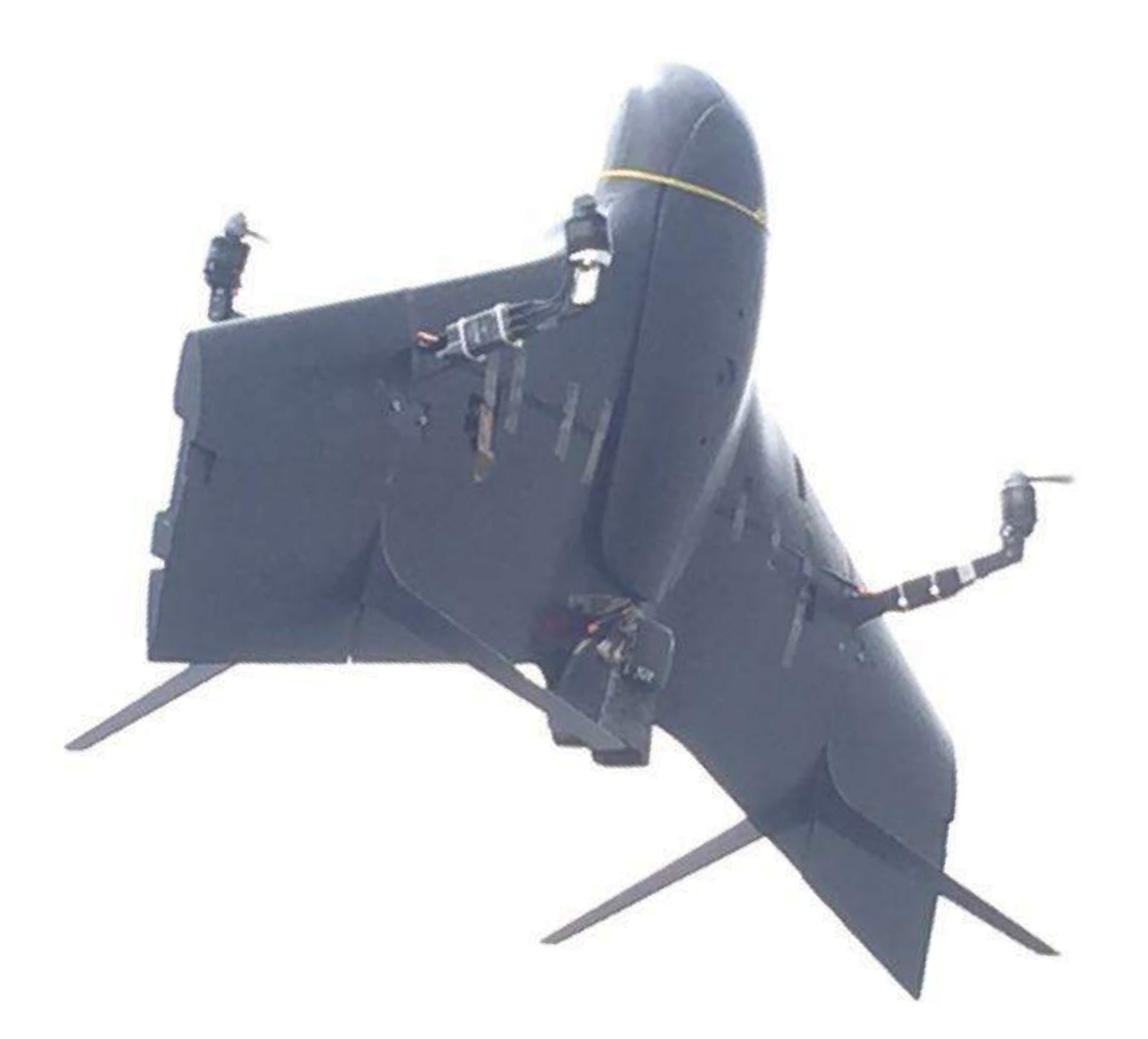}
		\caption{The tail-sitter VTOL UAV prototype used in this work\\
			(video link: https://youtu.be/xm8CZagg6V8)}
		\label{fig:tail_sitter_VTOL}
	\end{figure}
	
	In this paper, a unified control method is proposed for a tail-sitter VTOL UAV. Fig. 1 shows the tail-sitter aircraft used in this paper, which combines four rotors with a fixed-wing body. For details of the design and modeling process, interested readers are referred to \cite{zhang2017modeling} and \cite{lyu2017design}. The control method presented in this paper is unified, which means only one controller structure is used for all flight modes, regardless of whether it is in hovering, transition or level flight modes. Compared to the other transition control methods, which will be discussed in the next section, the advantage of the control strategy presented in this paper is that it can control the aircraft's full flight dynamics during the transition process. As it does not distinguish between hover and level flight mode, the transition is continuous and smooth and eliminates significant altitude drop or gain. 
	
	The essential part of this control method is to solve the attitude and collective rotor acceleration problem to achieve the required acceleration provided by the position controller in real-time. For this purpose, a nonlinear optimization problem based on an accurate UAV dynamics and aerodynamics model is formulated and solved by a nonlinear sequential convex programming (SCP) solver. To address the problem of high computational complexity of the nonlinear solver, a novel idea is employed, inspired by the widely-used imitation learning techniques in recent years. The idea is to apply long term short term memory (LTSM) to learn the behavior of the SCP solver, and then generate a neural network called SCPNet to replace the SCP solver for real-time implementation. The computation time for SCPNet is quite short, within 5 ms on an NVIDIA TX2 (an onboard computer used for real flights). Besides the low-computation cost, the proposed LSTM, with proper dynamics excitation and data collection during the training, can approximate the nonlinear solver solutions fairly accurately, as proved in the flight simulator \cite{zhang2017modeling}. After software-in-the-loop verification of SCPNet, we design several trajectories, including taking off, hovering, forward transition, level flight, backward transition, and landing, to demonstrate the effectiveness of this control method by real experiments in both indoor and outdoor environments. 
	
	Above all, the main contributions of this paper are:
	
	\begin{enumerate}
		\item We propose a unified controller design for a tail-sitter VTOL UAV, based on recurrent neural networks. An advantage of this design method is that the various flight modes (i.e., hovering, transition and level flight) of a VTOL UAV are controlled in a unified manner, as opposed to treating them separately and in the runtime switching one from another.
		\item We provide a new application of an RNN in UAV control, and that is to employ an RNN to approximate the behavior of a nonlinear solver with high computational complexity in the position controller loop, considering various engineering techniques, such as data collection, network training, and network deployment. 
		\item We perform systematic stability and robustness analysis on the proposed RNN-based controller by linearizing the closed-loop system around various flight speeds.    
		\item We conduct extensive simulation and experiments for all the flight modes (i.e., hovering, forward transition, level flight, back transition, and landing) both with and without disturbances to demonstrate the effectiveness of the method. 
	\end{enumerate}
	
	The remainder of this paper is organized as follows. In Section II, we discuss relevant literature. We introduce the system modeling of the tail-sitter UAV in Section III. The unified control method and imitation learning is described in section IV and V, respectively. Simulations and experiments results are analyzed in Section VI and VII, respectively. Finally, concluding remarks are provided in Section VIII.
	
	\section{Related Work}
	The first part of this section will present existing works related to flight control methods, and the second part of this section will focus on deep learning techniques for UAVs. Then, we propose a novel solution to solve the problem of high computational complexity using neural networks.
	
	There is a variety of literature on the flight control of tail-sitter VTOL UAVs. In \cite{stone2004control}, a T-wing tail-sitter aircraft is developed, and an LQR controller is applied to control both the forward and vertical flight. The transition flight is controlled by the vertical controller, which receives the pitch angle commands. However, an open-loop scheme is used to specify the throttle, leading to an altitude drop or gain during the transition.  In \cite{stone2001optimization}, a databased off-line optimization algorithm is applied to minimize the transition time, but it does not guarantee robustness and reliability when disturbances are present in the real operating environment. 
	
	Similarly, off-line optimization for transition control is presented in \cite{kita2010transition} and \cite{oosedo2017optimal}, and in \cite{frank2007hover}, Frank \textit{et al}.  use a set of trajectory points (relative to the pitch command) to realize the transition control; when the pitch angle is smaller than the set value, the hover controller will switch to the level flight controller. Three controllers are discussed on a Convair XFY-1 Pogo in \cite{knoebel2006preliminary}: a feedback linearization controller, vector-thrust model-based controller, and model reference adaptive controller (MRAC); however, all of these controllers are limited to two-dimensional flight dynamics, where only the longitudinal direction is controlled. More work on the tail-sitter aircraft's transition process control can be found in \cite{naldi2011optimal} and \cite{pucci2012flight}, but most of the work is limited to two-dimensions, and only simulation results are presented. 
	
	In \cite{naldi2011optimal}, the authors present two optimal problems for transition control, the objectives are minimum time and minimum energy, and the optimization problems are solved numerically and validated in simulation. Casau \textit{et al}. \cite{casau2013hybrid} develop a hybrid control method for the transition of a model-scale fixed-wing aircraft, for which there is no hover flight mode. The contribution of this paper is to divide the flight envelope into four regions: hover, transition, recovery and level flight, and different control techniques are used for each region.  The aerodynamic forces of a tail-sitter aircraft typically depend on the attitude (e.g., pitch angle). To make the design of the controller easier, the authors in \cite{pucci2011nonlinear} present a transformation method to make the aerodynamic forces independent of the attitude. Changing the control input of thrust enables the computation of attitude and thrust separately, thus easing the control design and analysis of the transition process. Furthermore, the authors in \cite{pucci2012flight} investigate the stall effect and find that the existence of control singularity and equilibrium bifurcation will cause some un-tractable maneuvers.
	
	Despite the research, transition control of tail-sitter VTOL UAVs is still a challenge. Most of the existing control methods apply multiple control frameworks for different flight modes, such as using a hover controller for hovering, a level flight controller for level flight, and a transition controller for the transition between hovering and level flight. However, the switching between different flight modes requires much tuning and often leads to an unsatisfactory transient response. In this paper, we present a unified control framework to control a tail-sitter VTOL UAV, building on our previous work in \cite{zhou2017unified}. In our prior work \cite{zhou2017unified}, the concept of unified control was proposed and proved in an ideal flight simulator. In this paper, we prove the effectiveness of the unified control method when all realistic flight characteristics, such as motor delay, saturation, gyroscopic effects, etc., are considered, based on the results in  \cite{zhou2017unified}. Stability and robustness analysis is carried out. Additionally, we propose a learning method to address the high computational complexity problem that occurred in our previous work. 
	
	Recent years have witnessed a rapidly growing trend of utilizing deep learning techniques for the control of UAVs. Travis \textit{et al}. \cite{dierks2010output} present a nonlinear feedback controller for a quadrotor UAV using a neural network to learn the complete dynamics of the UAV in an online fashion. The effectiveness of this control scheme is demonstrated in the presence of unknown nonlinear dynamics and disturbances. In \cite{nodland2013neural}, an optimal controller design method using a neural network (NN) is proposed for a helicopter UAV. A single NN is utilized for the cost function approximation to achieve optimal trajectory tracking, and finally, the overall closed-loop stability and effectiveness of trajectory tracking are demonstrated. In \cite{giusti2016machine}, Giusti \textit{et al}.  train a deep network to determine actions that can keep a quadrotor on a trail by learning on single monocular images collected from the robot's perspective. The optimal actions for the collected images can then be easily labeled. In \cite{hwangbo2017control}, Jemin \textit{et al}.  present a method to control a quadrotor UAV with an NN trained using reinforcement learning techniques. The trained network can derive the mapping of states and actuator commands directly, making any predefined control structure obsolete for training. The average computation time of this NN is about 7 $\mu$s per time step, rendering it more than efficient for the purpose of UAV control, for which the time scales are typically in the order of 10 ms.
	
	Although no prior works using a neural network to control tail-sitter VTOL UAVs have been found, the presented literature indicates the possibility of this application. Our task is to learn the behavior of the aforementioned SCP solver, which can be formulated as a typical imitation learning problem. Among the most widely-used imitation learning techniques, the task can be solved by behavior cloning. It solves problems in a supervised manner, by directly learning the mapping between the input observations and their corresponding actions, which are given by the expert policy. This formulation can give a satisfactory performance when sufficient training data is provided \cite{tai2016deep}. 
	
	\section{System Modeling}
	\subsection*{Nomenclature}
	
	\begin{supertabular}{ll}
		$\bm{p}$  & Position vector in the Earth frame\\
		$\bm{v}$  & Aircraft velocity vector in the Earth frame\\
		$m$ & Aircraft mass \\
		$\bm{f}_{t}$  & Force vector in the Earth frame\\
		$\bm{R}$ & Aircraft orientation w.r.t. the Earth frame\\ 
		$\bm{\omega}$  & Angular velocity in the body frame\\
		$\bm{I}$  & Inertia matrix of the aircraft\\
		$\bm{\tau}$  & Moment vector generated by the differential thrust\\
		$\bm{m}_{aero}$  & Aerodynamic moment in the body frame\\
		$\bm{f}_{aero}$  & Aerodynamic force in the body frame\\
		$T$  & Thrust produced by the four propellers\\
		$\bm{u}$ & Airspeed vector in the body frame\\
		$\bm{w}$ & Wind speed vector in the Earth frame\\
		$U$ & Airspeed magnitude\\
		$\alpha_x$ & Angle of attack\\
		$\beta$ & Sideslip angle\\
		$C_D$ & Drag coefficient\\
		$C_Y$ & Side force coefficient\\
		$C_L$ & Lift coefficient\\
		$C_l$ & Rolling momentum coefficient\\
		$C_m$ & Pitching momentum coefficient\\
		$C_n$ & Yawing momentum coefficient\\
		$S$ & Reference area\\    
	\end{supertabular}
	
	\subsection{Flight dynamics}
\begin{figure}[t!]
		\vspace{-0cm}
		\centering
		\includegraphics[width=2.6in]{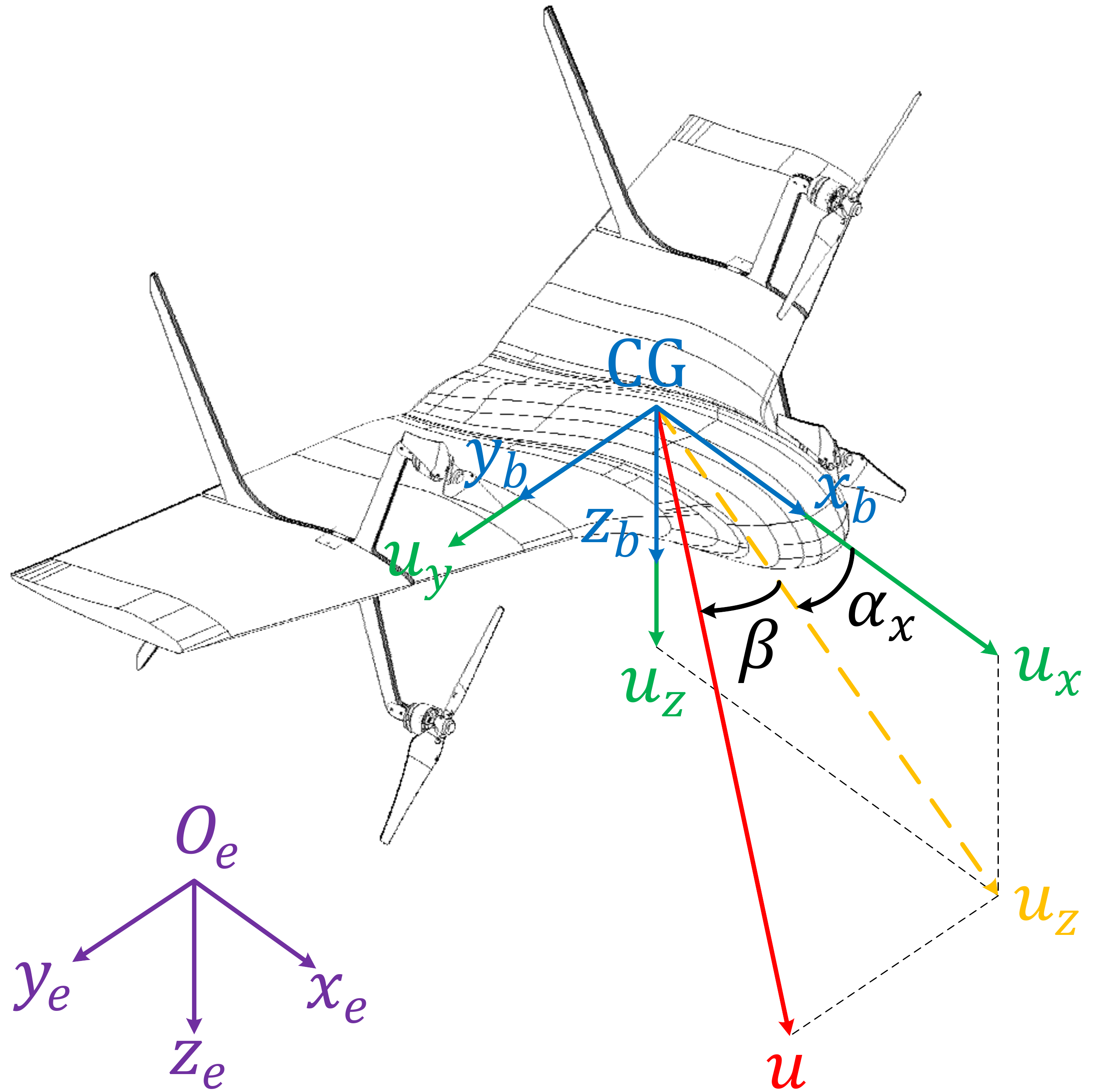}
		\caption{Earth frame versus body frame}
		\label{f:taisitter_frame}
		\vspace{-0cm}
	\end{figure}

	In this paper, the coordinate systems of conventional fixed-wing aircraft are used \cite{nelson1998flight, etkin1959dynamics}. Figure \ref{f:taisitter_frame} shows the Earth frame $F_e(O_e, X_e, Y_e, Z_e)$ in the convention of north, east and down, and the body frame $F_b (CG, X_b, Y_b, Z_b)$ in terms of the front, right and down direction of the aircraft, and $ \bm{R} \in SO(3)$ is defined as the orientation of the body frame w.r.t. the Earth frame. Applying Newton's equations of motion to the airframe yields the following dynamic model:
	
	\begin{subequations}
		\begin{eqnarray}
		\label{e:p_dot}
		\dot {\bm{p}}& =&  \bm{v}  \\
		\label{e:trans_eqn}
		m \dot{\bm{v}}& =& \bm{f}_t \\
		\dot{\bm{R}} &=& \bm{R} \widehat{\bm{\omega}}\\
		\label{e:omg_dot}
		{\bm{I}} \dot{\bm{\omega}}   &=& - \bm{\omega} \times (\bm{I} \bm{\omega} )  + \bm{\tau}  + \bm{m}_{aero}, 
		\end{eqnarray}
	\end{subequations}
	where $\bm{p} = [x,y,z ]^T $ is the position of the aircraft's mass center expressed in the Earth frame, $\bm{v} = [v_x, v_y, v_z]^T $ is the aircraft's linear velocity expressed in the Earth frame, $\bm{f}_t$ is the total force applied to the aircraft,  $\bm{\omega}$ denotes the angular rate of the aircraft expressed in the body frame, $m$ is the total mass of the aircraft, $\bm{I}$ is the aircraft's inertia matrix, $\bm{\tau}$ denotes the moment vector generated by the differential thrust, and $\bm{m}_{aero}$  is the aerodynamic moment expressed in the body frame.  $\widehat{\bm{\omega}}$ is a skew-symmetric matrix, and $\widehat{\bm{\omega }} \bm{x} = \bm{\omega} \times \bm{x} $ holds for any vector $\bm{x} \in \mathbb{\bm{R}}^3$.
	
	\begin{figure}[t!]
		\vspace{0cm}
		\centering
		\includegraphics[width=2in]{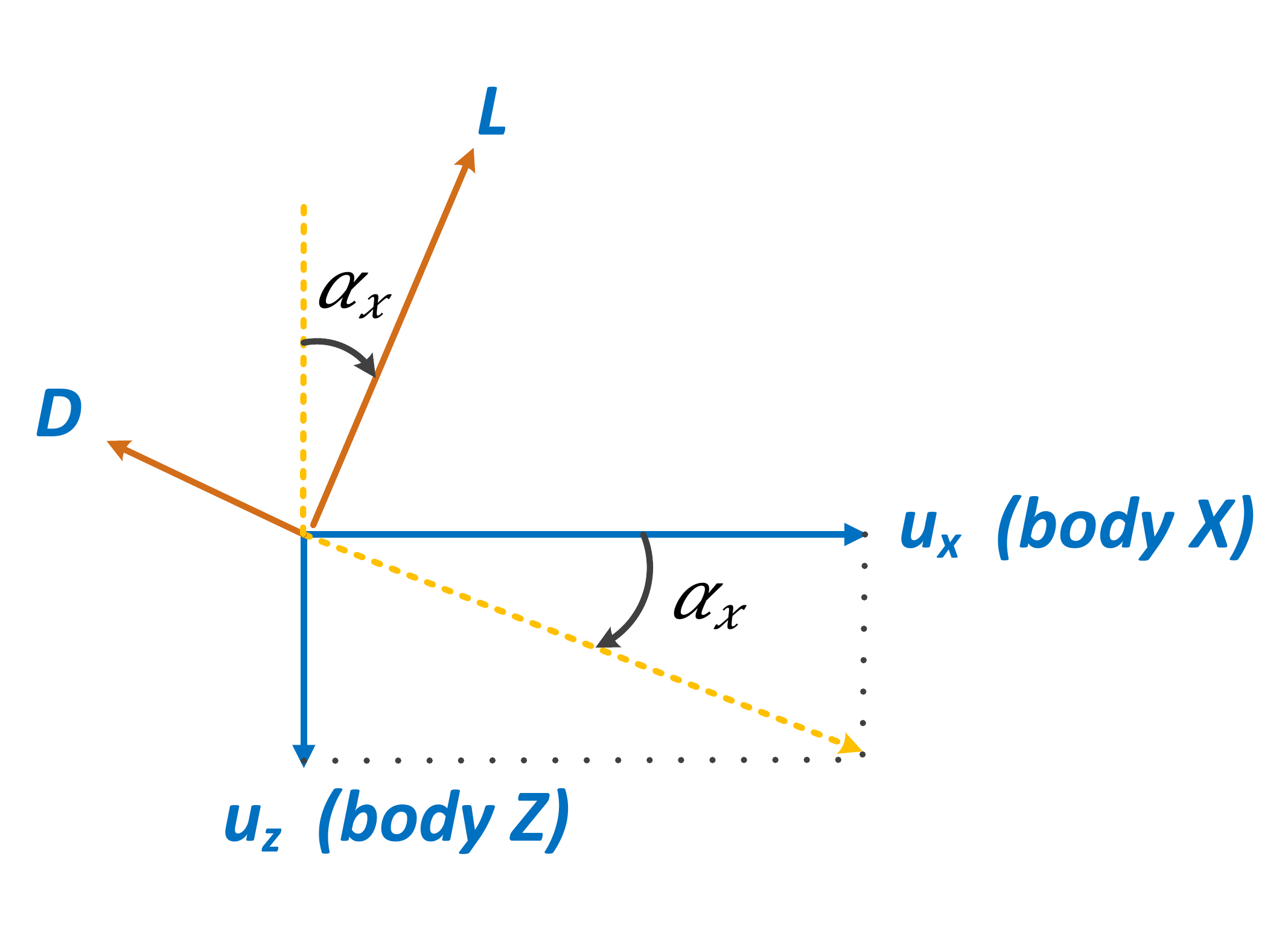}
		\caption{Lift and drag}
		\label{fig:lift_drag}
		\vspace{-0cm}
	\end{figure}
	
	Forces applied on the vehicle $\bm{f}_t$ include gravity, propeller thrust and aerodynamic force, and it can be expressed as
	\begin{equation}
	\label{e:f_t}
	\bm{f}_t = \left[ \begin{array}{c} 
	0  \\  0 \\ mg
	\end{array} \right] +
	\bm{R} \left ( 
	\left[ \begin{array}{c} 
	T \\ 0  \\  0
	\end{array} \right]
	+  \bm{f}_{aero} \right ),
	\end{equation} 
	where $T$ is the total thrust produced by the 4 propellers, and $\bm{f}_{aero}$ is the aerodynamic force, which are represented in the body frame, while $\bm{f}_t$ is represented in the Earth frame. 
	
	\subsection{Aerodynamic modeling}
	Beginning with the analysis of a tail-sitter aircraft's aerodynamics, the airspeed $\bm{u}$ should be defined as:
	\begin{eqnarray}
	\label{e:airspeed}
	\bm{u} = \bm{R}^T (\bm{v} - \bm{w}),
	\end{eqnarray}
	where $\bm{v}$ denotes the linear velocity of the aircraft in Eq. (\ref{e:p_dot}) and $\bm{w}$ refers to the wind speed, which can be estimated by wind estimation algorithms \cite{cho2011wind} or measured by sensors such as a pitot tube. The magnitude of airspeed $\bm{u}$, angle of attack $\alpha_x$ and  angle of sideslip $\beta$ are respectively expressed as
	
	\begin{subequations}
		\begin{eqnarray}
		\label{e:airspeed_mag}
		U & = & \sqrt{u_x^2 + u_y^2 + u_z^2} \\
		\label{e:alpha}
		\alpha_x & = & \tan^{-1} \left( \frac{u_z}{u_x} \right) \\
		\beta & = & \sin^{-1} \left( \frac{u_y}{U} \right).
		\end{eqnarray}
	\end{subequations}
	
	\begin{figure}[t!]
		\centering
		\includegraphics[width=2.5in]{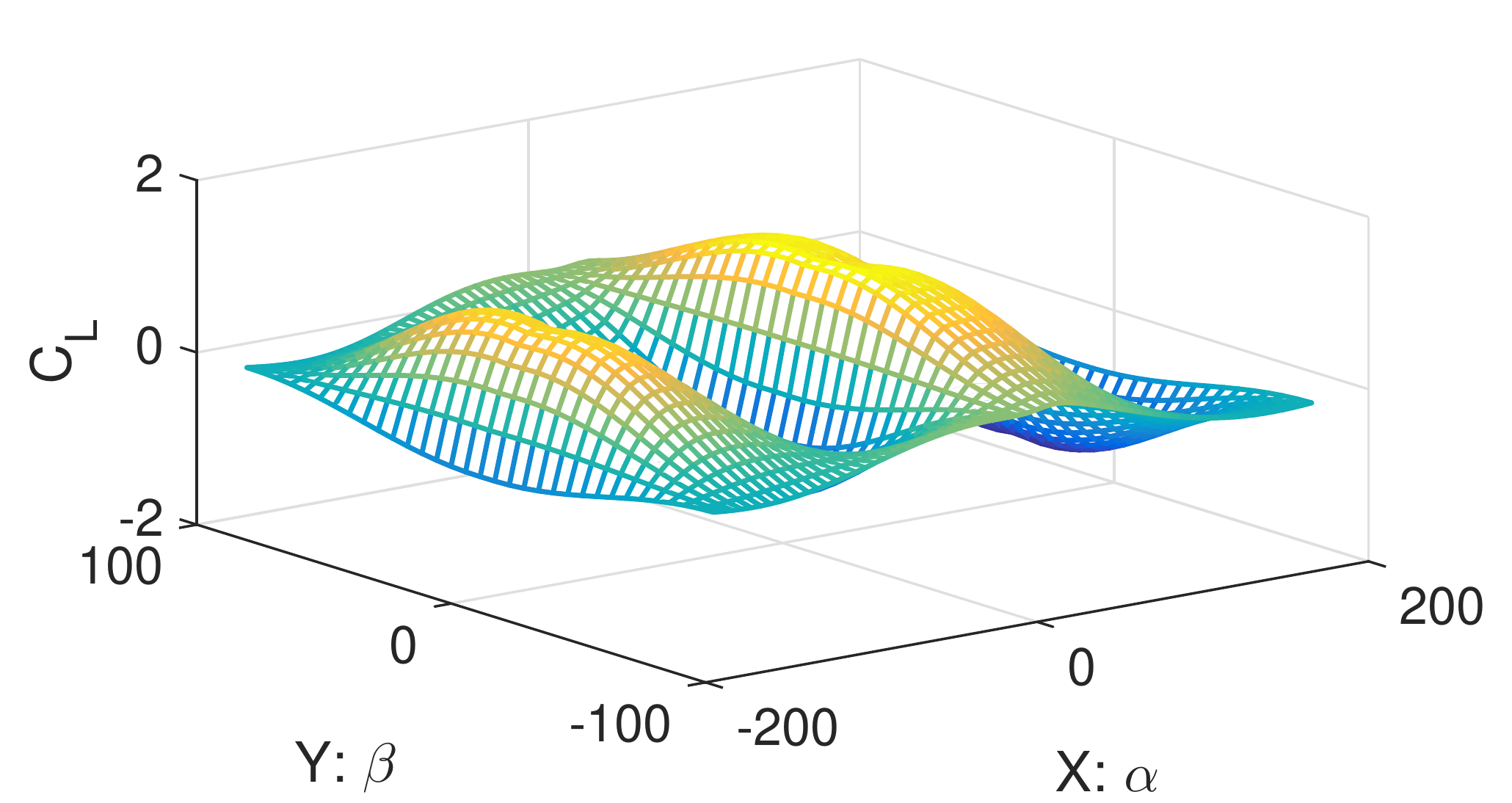}
		\caption{Full envelope lift coefficient $C_L$}
		\vspace{-0.2cm}
		\label{fig:lift_coeff}
		\vspace{-0.1cm}
	\end{figure}
	
	The aerodynamic forces are expressed as $\bm{f}_{aero} = \left[ X, Y, Z\right]^T$ and  the aerodynamic moments are expressed as $\bm{m}_{aero} = \left[ l, m, n\right]^T$ in the body frame. Generally, the aerodynamic forces are discussed in the stability axes frame \cite{hamel2002dynamic} where lift $L$, drag $D$ and side force $Y$ are included. Fig. \ref{fig:lift_drag} shows that the lift is perpendicular to the airspeed, and drag is in the opposite direction of the airspeed. The right-handed frame of the drag, side force and lift follow the definitions
	\begin{eqnarray}
	\label{e:f_xyz}
	\left[ 
	\begin{array}{c}
	X \\ Y \\ Z
	\end{array}
	\right] = 
	\left[ 
	\begin{array}{ccc}
	-\cos \alpha_x & 0 & \sin \alpha_x \\
	0 & 1 & 0 \\
	-\sin \alpha_x & 0 & - \cos \alpha_x
	\end{array}
	\right]
	\left[ 
	\begin{array}{c}
	D \\ Y \\ L
	\end{array}
	\right],
	\end{eqnarray}
	and the lift $L$, drag $D$, side force $Y$ and moments are
	\begin{eqnarray}
	\begin{array}{ccc}
	L = C_L Q S; & D = C_D Q S; & Y = C_Y Q S \\
	l = C_l Q S \bar{c}; & m = C_m Q S \bar{c}; & n = C_n Q S \bar{c},
	\end{array}
	\nonumber
	\end{eqnarray}
	where $C_L$, $C_D$ and $C_Y$ denotes the coefficients of the lift force, drag force and side force, respectively; $C_l$, $C_m$ and $C_n$ represents the coefficient of the rolling moment, pitching moment, and yawing moment, respectively; $Q = \frac{1}{2} \rho U^2 $ denotes the dynamic pressure, $S$ denotes the area; and $\bar{c}$ denotes the mean aerodynamic chord (MAC). For a tail-sitter aircraft whose operating speed is far from the speed of sound, the coefficients depend on the angle of attack $\alpha_x$ and angle of sideslip $\beta$, which can be derived from wind tunnel experiments. For compactness, we only show the full envelope of the lift in Fig. \ref{fig:lift_coeff}. The drag, side force coefficients, and the detailed wind tunnel experiments can be found in \cite{zhang2017modeling}.    
	
	With the models of the flight dynamics and aerodynamics at hand, the control design can be carried out. Detailed procedures will be shown in the next section.
	
	\section{Unified Control Method}
\begin{figure}[t]
		\vspace{0.1cm}
		\centering
		\includegraphics[width=3.5in]{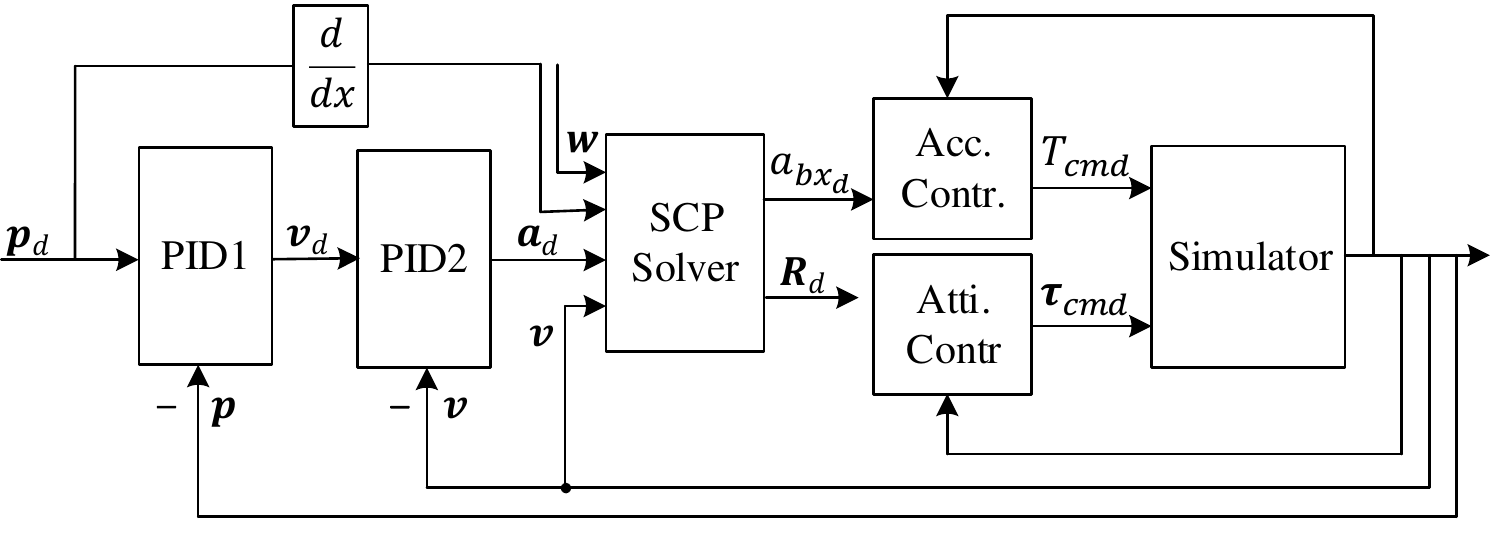}
		\caption{Control structure}
		\label{fig:contrl_structure}
		\vspace{-0cm}
	\end{figure}

	The proposed unified control structure is shown in Fig. \ref{fig:contrl_structure}, which consists of an outer loop position controller providing the desired attitude and body X acceleration and two inner loop controllers tracking each of the desired acceleration and attitude. In this section, we will introduce the unified control framework in detail, except for the attitude controller part, for which interested readers can refer to \cite{zhou2017unified} and \cite{zhou2018frequency}. 
	
	\subsection{Underlying acceleration control}    
	Based on the dual-loop control structure, an acceleration controller along the X direction of the body frame is employed to track the desired body X acceleration computed by the position controller by adjusting the aircraft throttle. In the designed VTOL UAV system, we set the throttle command in the range of [0, 1], which is mapped to the collective thrusts from the four rotors. The feedback of the acceleration controller is from the onboard IMU, which provides high-rate acceleration measurements in all the three body axes. In our case, only the component along the body X axis is used.  Based on these, a  PID controller is applied to track the body X acceleration command $a_{bx_d}$ generated by the position controller, shown as the following equation:
	\begin{eqnarray}
	T_{cmd} = K_{p} a_e + K_{i} \int a_e dt + K_{d} \frac{d a_e}{dt},
	\end{eqnarray}
	where $T_{cmd}$ is the throttle command sent to the actuators, $a_e$ is the tracking error of the body X acceleration, and $K_{p}$, $K_{i}$ and $K_{d}$ is the proportional gain, integral gain, and derivative gain, respectively, for the PID controller in the acceleration loop. 
	
	\subsection{Unified position control based on the SCP solver}
	Combining Eq. (\ref{e:p_dot}), Eq. (\ref{e:trans_eqn}) and Eq. (\ref{e:f_t}), the translational motion of the aircraft can be rewritten as 
	\begin{subequations}
		\begin{eqnarray}
		\label{e:tran_dynamics}
		\dot {\bm{p}} & = & \bm{v} \\
		m \dot{\bm{v}} & = & m g \bm{e}_3 + \bm{R} \left(  T \bm{e}_1 + \bm{f}_{aero}\right),
		\label{e:dynamics}   
		\end{eqnarray}
	\end{subequations}
	where $T$ is the total thrust generated by the 4 propellers, $\bm{e}_1 = [1,0, 0]^T$, and $\bm{e}_3 = [0,0, 1]^T$. For a pre-set trajectory $\bm{p}_d(t)$, the position error can be computed as $\bm{e}_p = \bm{p}_d(t) - \bm{p}(t)$. Then, a cascaded control architecture consists of an outer position controller and an inner velocity controller is applied to compute the desired acceleration, shown as below:
	\begin{subequations}
		\begin{eqnarray}
		\label{e:vel}
		\bm{v}_d = \dot{\bm{p}}_d(t) + K_{p_1} \bm{e}_p + K_{i_1} \int \bm{e}_p d t + K_{d_1} \frac{d \bm{e}_p}{dt} \\
		\label{e:pos}
		\bm{a}_d = \dot{\bm{v}}_d(t) + K_{p_2} \bm{e}_v + K_{i_2} \int \bm{e}_v d t + K_{d_2} \frac{d \bm{e}_v}{dt},
		\end{eqnarray}
	\end{subequations}
	where $\bm{v}_d$ is the desired velocity, and $\bm{a}_d$ is the desired acceleration, and each of these is computed by a PID controller with a feed forward term.
	
	To achieve the desired acceleration $\bm{a}_d$, the desired attitude $\bm{R}_d$ and body X acceleration $a_{bx_d}$ in Fig. \ref{fig:contrl_structure} should satisfy the following equation: 
	
	\begin{eqnarray}
	\label{e:accel_eq_new}
	g \bm{e}_3 + \bm{R_d}\bm{e}_1 a_{bx_d} + \frac{1}{m} \bm{R_d} \bm{f}_{aero} = \bm{a}_d,
	\end{eqnarray}
	which is derived by dividing both sides of Eq. (\ref{e:dynamics}) by $m$, and substituting $\frac{1}{m} a_{bx_d}$ and $\bm{a}_d$ for $T$ and $\dot{\bm{v}}$, respectively.
	
	In the controller design of conventional quad-rotors, the aerodynamic force $\bm{f}_{aero}$ is usually ignored, and the analytical solution of the attitude and body X acceleration in Eq. (\ref{e:accel_eq_new}) can be derived \cite{hamel2002dynamic} easily. However, a tail-sitter aircraft uses the aerodynamic force to provide lift during a level flight to improve efficiency. Thus, the aerodynamic forces cannot be ignored when computing the attitude and body X acceleration. 
	
	As discussed in Section III, the aerodynamic forces are related to the airspeed $\bm{u}$ and aerodynamic coefficients, while the latter variables depend on the angle of attack $\alpha_x$ and angle of sideslip $\beta$, which can be calculated from the airspeed $\bm{u}$. Therefore, we describe the aerodynamic forces as a nonlinear function of the airspeed $\bm{u}$. Substituting $\bm{u}$ into Eq. (\ref{e:airspeed}), we obtain the following equation:
	\begin{eqnarray}
	\bm{f}_{aero} = f(\bm{R}_d^T(\bm{v} - \bm{w})).
	\end{eqnarray}
	
	The dependence of the aerodynamic force in Eq. (\ref{e:accel_eq_new}) on the attitude makes the equation extremely nonlinear and difficult to solve. In our prior work \cite{zhou2017unified}, we employ a numerical approach to solve for the proper attitude and body X acceleration for a given inertial acceleration, inertial velocity, and wind speed. More specifically, we formulate an optimization problem with equality constraints from Eq. (\ref{e:accel_eq_new}). The objective function attempts to align the vehicle's heading with the direction of velocity $\bm{v}$, which is consistent with the aerodynamic requirements and the actual flight experience. As a result, the optimization problem can be constructed as
	\begin{equation}
	\begin{split}
	\min_{\left[\bm{\xi}, T \right] \in \mathbb{R}^4} & \ \  \ | v_{by} | \\ 
	\text{s.t.} &  \ \  g \bm{e}_3 + \bm{R}_d\bm{e}_1 a_{bx_d} + \frac{1}{m} \bm{R}_d f(\bm{R}_d^T (\bm{v} - \bm{w})) - \bm{a}_d = 0 \\ 
	& \ \ \bm{R}_d = \bm{R}_z \bm{R}_x \bm{R}_y \\
	& \ \ \min {a_{bx_d}}\leq {a_{bx_d}} \leq \max {a_{bx_d}},
	\end{split}
	\label{e:opt_prob}
	\end{equation}
	where $v_{by}$ is the body velocity in the Y direction. In the objective function,  $v_{by}$ is formulated to be minimized as we want the velocity along the Y direction to be as small as possible, which is reasonable to flying operation experience. In this optimization problem, we parameterize the attitude $\bm{R}$ by the ZXY Euler angle representation as the singular point is rarely reached for tail-sitter aircraft: more specific reasons were mentioned previously in the modeling section. The desired body X acceleration $a_{bx_d}$ should be bounded by its limits $\min {a_{bx_d}}$ and $\max {a_{bx_d}}$, which depend on the maximum and minimum thrust generated by the 4 motors. In the controller design, we also consider the saturation conditions: When $a_{bx_d}$ is outside the range of  ${a_{bx_d}}_{\min}$ and ${a_{bx_d}}_{\max}$, the integrator parameter will be set to zero. 
	
	In our previous work \cite{zhou2017unified}, a sequential convex programming (SCP) algorithm is employed to solve the non-convex optimization problem in Eq. (\ref{e:opt_prob}). It is an iterative method that solves non-convex optimization problems by iteratively solving a sequence of convex optimization sub-problems that approximate the cost and constraints of the original problem \cite{schulman2014motion}. This SCP algorithm is implemented using a MATLAB package, CVX, and the parameters are appropriately tuned for Eq. (\ref{e:opt_prob}). The attitude and body X acceleration derived from this equation will achieve the required acceleration $a_d$ to eliminate the position error. Therefore, the vehicle can be uniformly controlled in hovering, transition and level flight, regardless of the flight modes.    
	
	To verify the effectiveness of the proposed method, a simulator is established \cite{zhang2017modeling} considering various realistic issues to make the simulation performance close to a real system. After this,  several typical trajectories are tested to verify the unified control method. The results show good tracking performance and perfect transition process, showing a significant advantage over other control methods (discussed in Section II in related work), especially for the transition process, where the altitude drop or gain is less than 0.4 m. Readers can refer to \cite{zhou2017unified} for more details on the robustness, stability, and performance evaluation of the SCP methods. 
	
	In this section, we introduce a unified controller method based on the SCP solver proposed in our prior work. For this problem setup, the computational time of the solver implemented on an NVIDIA TX2 (an onboard computer used for real flight) is 1 s on average, and the maximum computational time will exceed 10 s in extreme cases where there exist large disturbances. The position controller, on the other hand, runs at the frequency of 50 Hz and requires the solver to converge within 0.02 s. To resolve this computational issue, we propose in the next section a recurrent neural network (RNN) to approximate the behavior of the SCP solver, thereby making the real-time implementation of such a unified controller plausible.

	\section{Imitation learning of the SCP solver}
	This section focuses on the pipeline to build SCPNet, where a neural network model is used to mimic the behavior of the SCP solver through imitation learning. We first briefly introduce the concept and algorithms of imitation learning, and later, the details of data collection, network structure, training, and testing for our experimental implementation are introduced.
	
	\subsection{Imitation learning}
	Unlike reinforcement learning where the policy is searched through a hand-designed reward, imitation learning is an alternative to learning control policies by providing the learning agent with experts' demonstrations\cite{bagnell2015invitation}. It offers a paradigm for agents to learn successful policies in fields where people can easily demonstrate the desired behavior but find it difficult to hand program or hardcode the correct cost or reward function. This is especially useful for humanoid robots or manipulators with high degrees of freedom. For our SCP solver, the behavior of outputs according to different inputs can be easily demonstrated via the simulator. When the learning accuracy reaches a high level, the newly-generated SCPNet learning from the SCP solver can substitute the SCP solver for real-time implementation. 
	
	\begin{figure}[t!]
		\centering
		\includegraphics[width=3.4in]{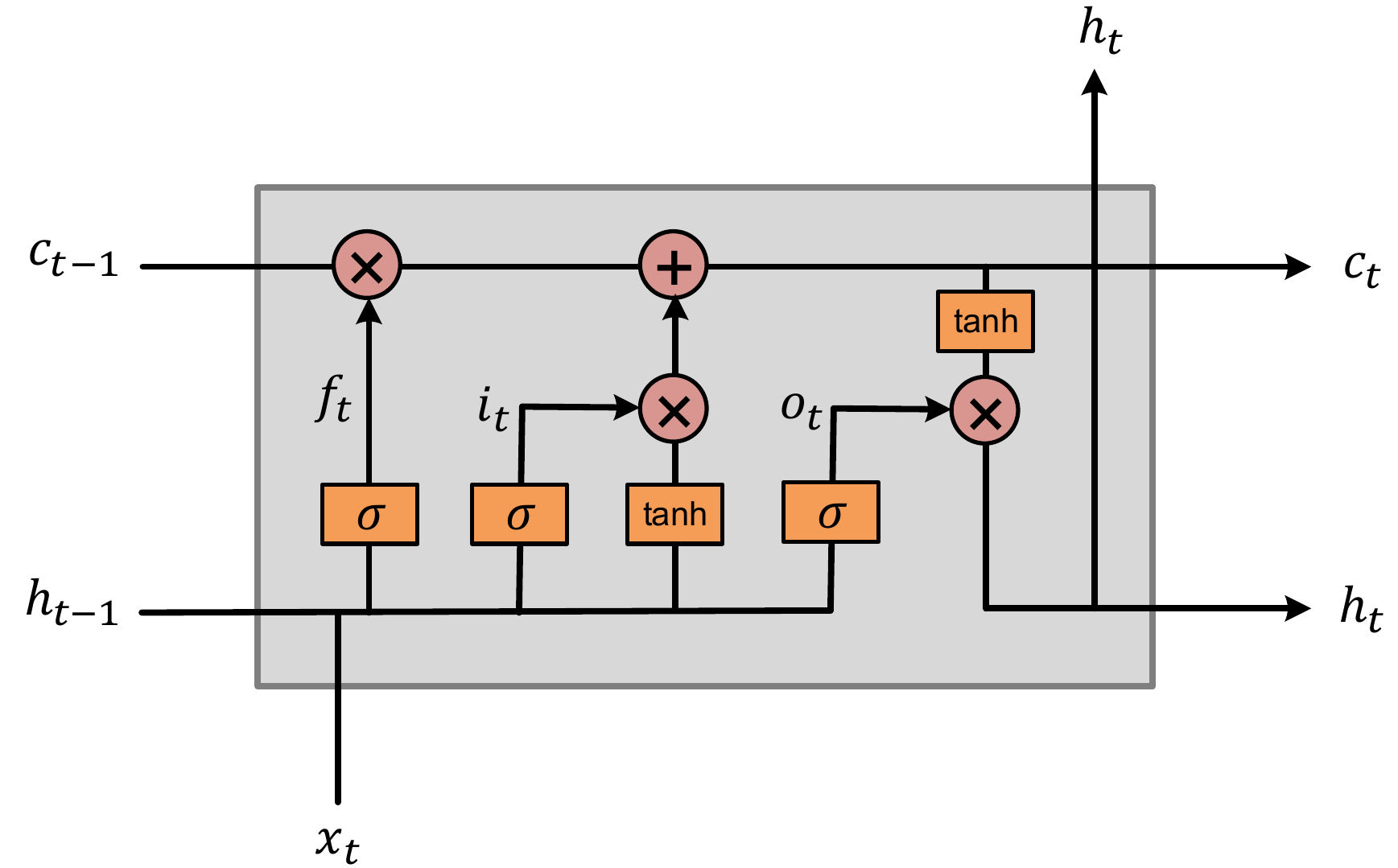}
		\caption{Long short-term memory cell}
		\label{fig:lstm}
		\vspace{-0cm}
	\end{figure}
	
	\begin{figure*}[thpb]
		\centering
		\includegraphics[width=0.9\textwidth]{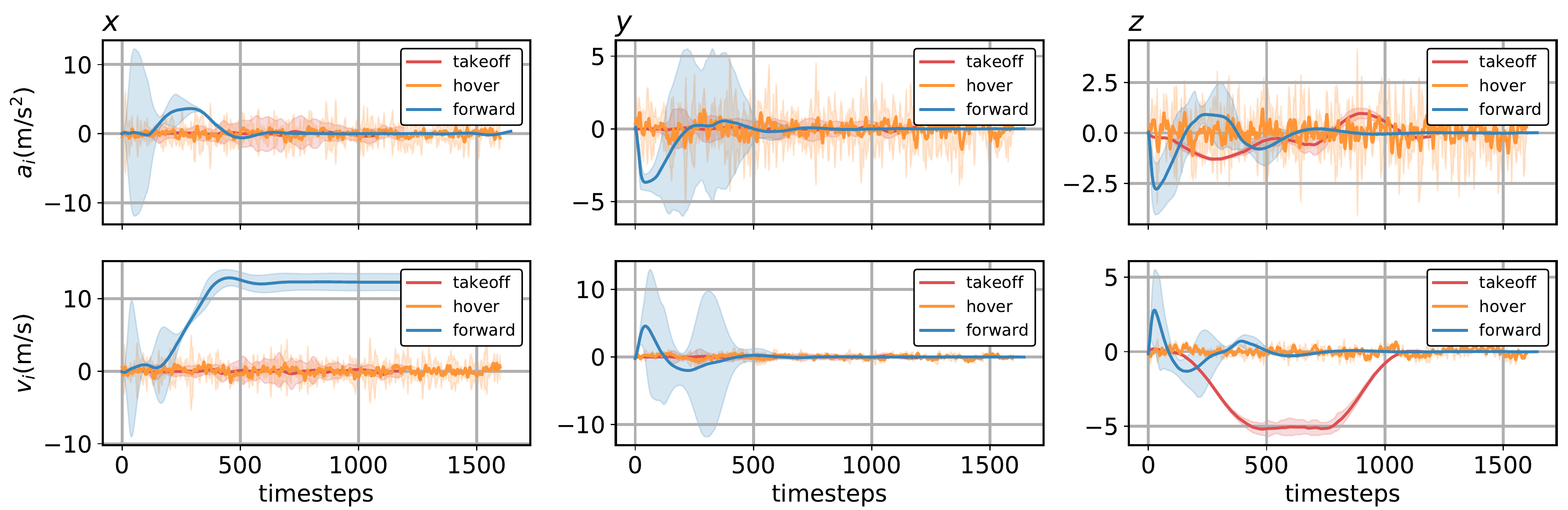}
		\caption{Training data for the neural network: inputs of inertial acceleration (above) and inertial velocity (below)
			\label{fig:data_training}}
	\end{figure*}
	
	In this paper, we use a neural network model to learn the behavior of the SCP solver. Besides the current inputs, the solver also needs to consider historical states to compute the appropriate outputs. Therefore, we apply an RNN structure for learning, which targets sequential prediction problems \cite{nguyen2016imitation}. These structures are widely used to tackle supervised problems involving learning a mapping from an input sequence to an output sequence.   
	
	Hochreiter \& Schmidhuber \cite{hochreiter1997long} proposed a special type of RNNs called long short-term memory (LSTM), which applies purpose-built memory cells to store information, thus mitigating the long-term dependence problem of conventional RNNs. A single LSTM memory cell is shown in Fig. \ref{fig:lstm}. In this paper, we use the standard version of LSTM \cite{gers2002learning}, which is expressed by the following function:
	\begin{center}
		\begin{subequations}
			\begin{eqnarray}
			i_t & = & \sigma \left(W_{xi}x_t + W_{hi}h_{t-1} + W_{ci}c_{t-1} + b_i\right) \\
			f_t & = & \sigma \left(W_{xf}x_t + W_{hf}h_{t-1} + W_{cf}c_{t-1} + b_f\right) \\
			c_t & = & f_tc_{t-1} + i_t{\rm tanh}\left(W_{xc}x_t + W_{hc}h_{t-1} + b_c\right)\\
			o_t & = & \sigma \left(W_{xo}x_t + W_{ho}h_{t-1} + W_{co}c_{t} + b_o\right) \\
			h_t & = & o_t{\rm tanh}\left(c_t\right),
			\end{eqnarray}
		\end{subequations}
	\end{center}
	where $\sigma$ is the sigmoid function, $i$, $f$, $o$ and $c$
	is the \textit{input gate}, \textit{forget gate}, \textit{output gate} and \textit{cell} activation vector, respectively, all four vectors have the same length as the hidden vector $h$, and the $W$ terms represent weight matrices (e.g., $W_{ci}$ denotes the cell-gate weight matrix).
	
	In our design, we use a sliding-window-like LSTM structure, where the current LSTM cell only depends on the last 5 historic cell outputs.  
	
	\begin{figure}[t!]
		\centering
		\includegraphics[width=2.1in]{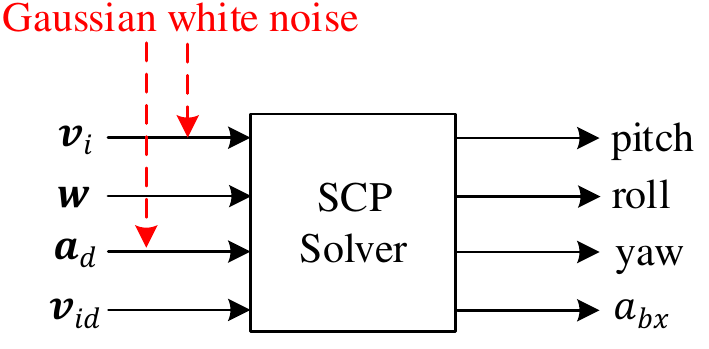}
		\caption{SCP solver structure}
		\label{fig:data_disturbance}
		\vspace{-0cm}
	\end{figure}
	
	\begin{figure}[t!]
		\centering
		\includegraphics[width=2.5in]{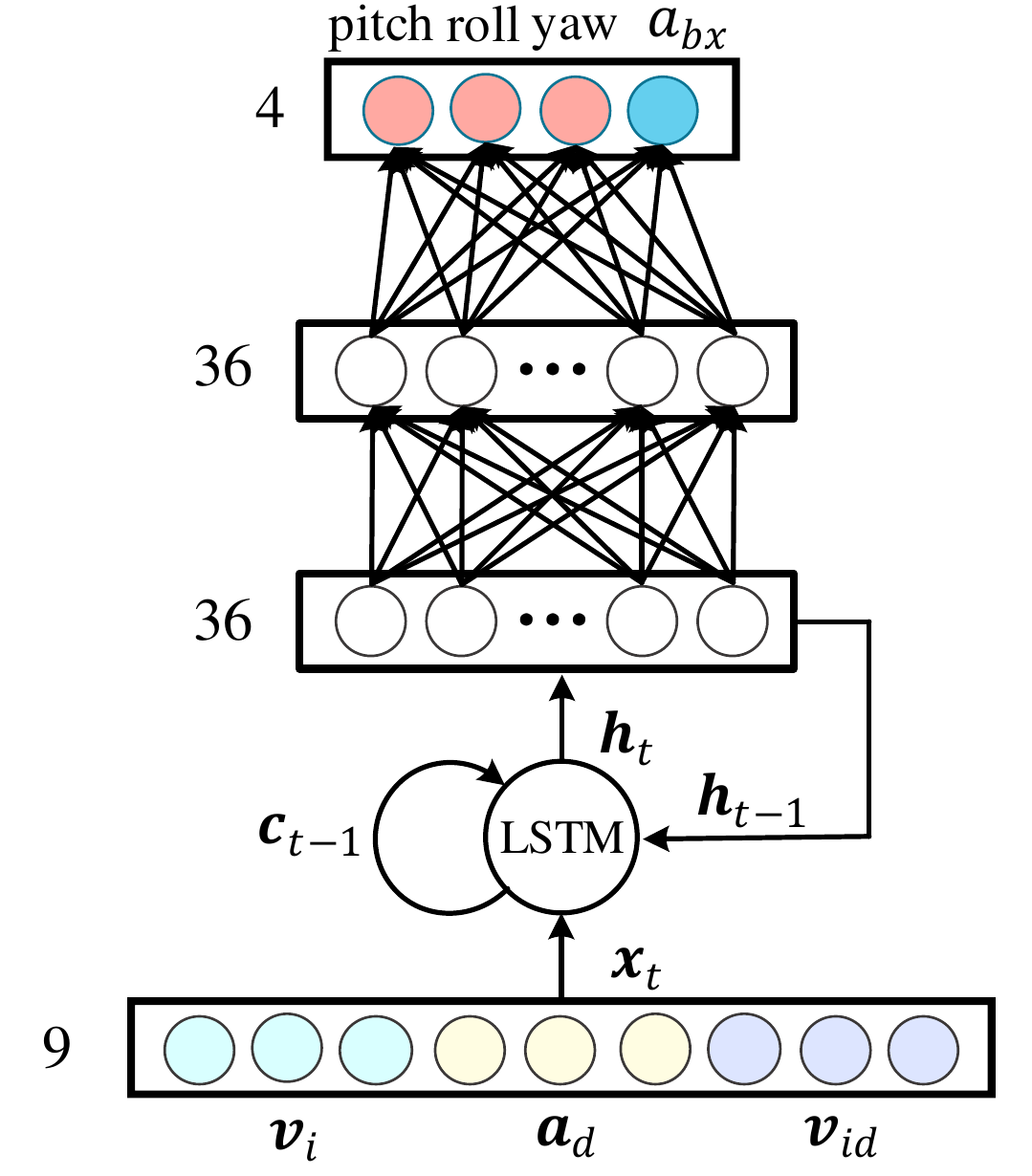}
		\caption{SCPNet structure}
		\label{fig:network}
		\vspace{-0cm}
	\end{figure}

	\subsection{Data collection}
	
	To excite the possible states of the aircraft near specified trajectories such that the trained SCPNet is robust to disturbances, we add Gaussian white noise disturbance to the inputs of the SCP solver when collecting data for training. The noise is added to the actual inertial velocity $v_i$ and the desired acceleration $a_d$, shown in Fig. \ref{fig:data_disturbance}. The input and output data of the SCP solver are generated through the predefined control framework in the closed-loop simulator shown in Fig. \ref{fig:contrl_structure}, without the requirement of manual label work. It can be regarded as a self-supervised learning approach.
	
	In this paper, we collect training data for the following three types of trajectories: takeoff, hovering and forward. Here the forward trajectories include the transition process and forward flight. The training data has 400 sequences for each kind of trajectory, and the number of timesteps of the sequence for takeoff, hovering and forward trajectory is 1200, 1600, and 1650, respectively. Fig. \ref{fig:data_training} shows parts of the training data. The three figures in the top row show the distribution of the inertial accelerations along each of the X, Y, and Z axes, and the figures in the bottom row show the distribution of inertial velocities. Each color represents one kind of trajectory, the lines represent the mean values, and the color-filled areas represent the variance fluctuations near the mean. From the figures, we can find the inertial acceleration and velocity ranges for each direction, which are reasonable for real flights, e.g., the mean value for forward flight is 12 m/s, which is the optimal cruising speed for this VTOL aircraft. Additionally, the data for other inputs and outputs are also vibrated in some ranges, as the data collection process is conducted in a closed-loop way.  
	
	Finally, all the 1200 sequences are randomly divided into two parts: training sequences and test sequences, with a ratio of 9:1. As a result, a total number of 1080 sequences are used for training and 120 sequences are used for testing. The takeoff, hovering, and forward trajectories are randomly distributed in both the training and testing sequences.    
	
	\begin{figure*}[t!]
		\centering
		\includegraphics[width=0.98\textwidth]{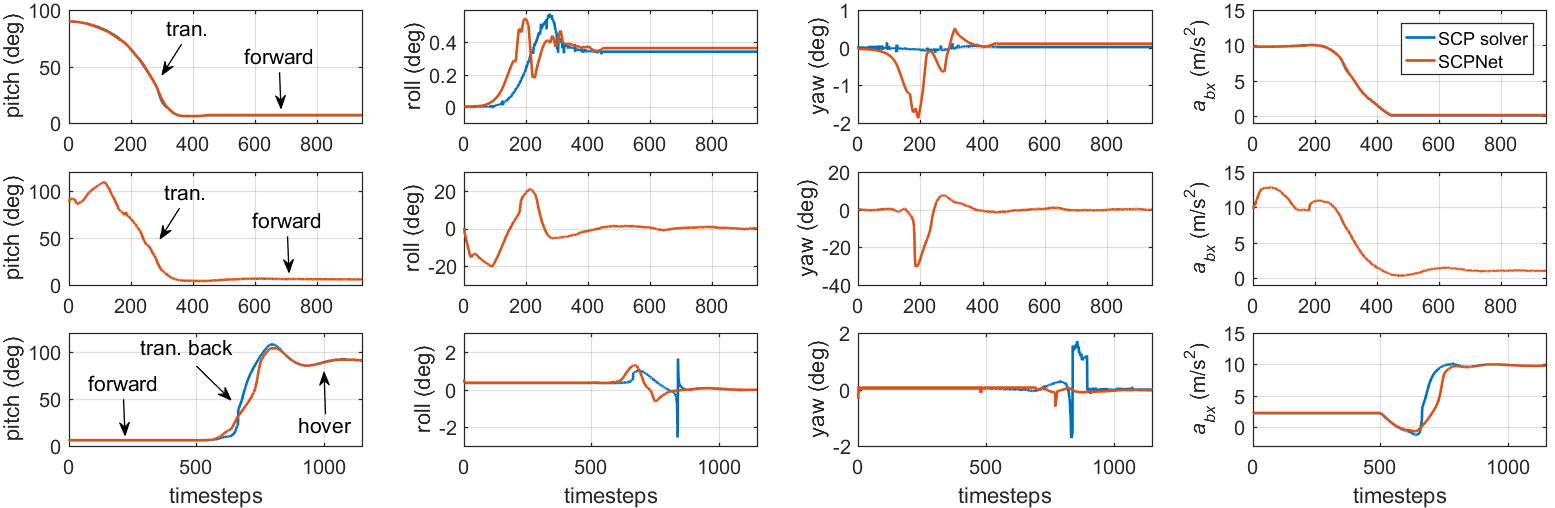}
		\caption{Test results for SCPNet: transition and forward flight with no disturbance (top); transition and forward flight with disturbance (middle); forward flight and transition back with no disturbance (bottom)}
		\label{fig:scp_test}
		\vspace{-0cm}
	\end{figure*}
	
	\subsection{Network structure}
	The network structure of SCPNet is shown in Figure \ref{fig:network}. The inputs of this network have 9 dimensions, which are the current velocity, desired acceleration, and the desired velocity. They are directly fed to the LSTM layer, which outputs a vector of 36 dimensions, then, the LSTM outputs are fed to a fully connected (FC) network. The FC network has one ReLU hidden layer with a dimension of 36 and a regression output layer to predict the attitude (i.e., pitch, roll, and yaw) and the body X acceleration. 
	
	\begin{figure}[t!]
		\centering
		\includegraphics[width=3.2in]{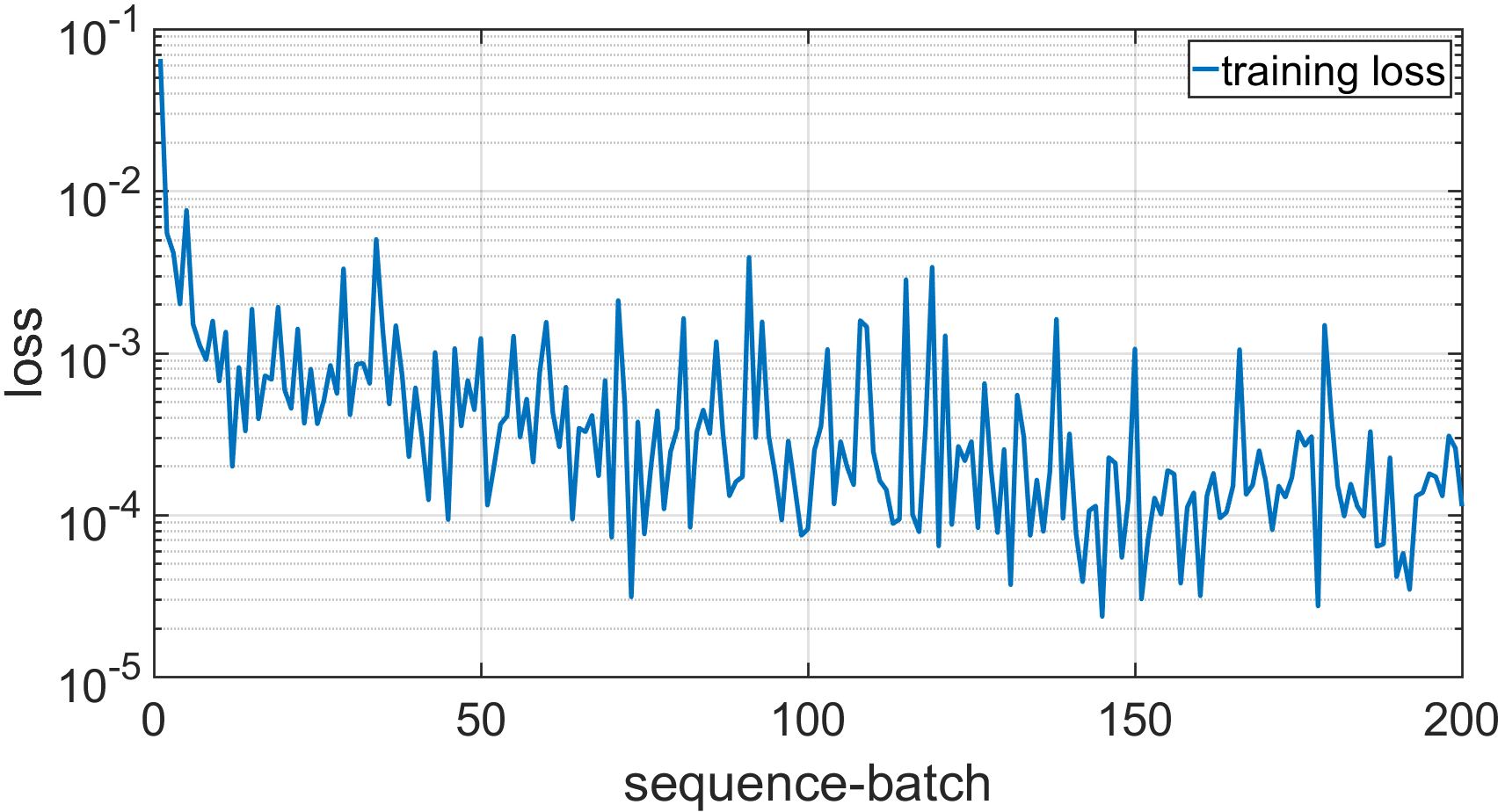}
		\caption{Training loss}
		\label{fig:TrainningLoss}
		\vspace{-0cm}
	\end{figure}
	
	\begin{figure}[t!]
		\centering
		\includegraphics[width=3.5in]{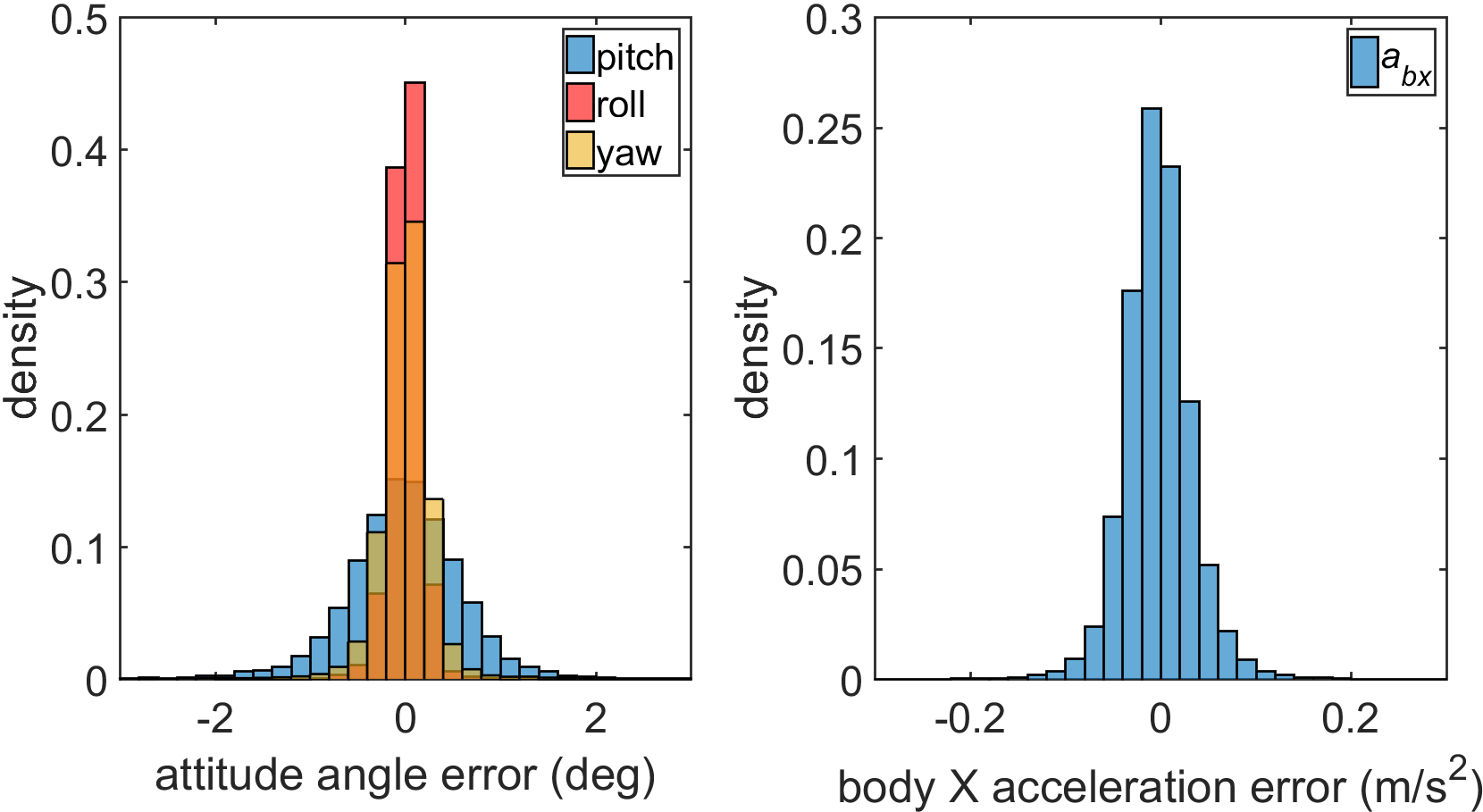}
		\caption{Error distribution of attitude and body X acceleration}
		\label{fig:ErrorAngleAbx}
		\vspace{-0cm}
	\end{figure}
	
	\subsection{Network training}
	To train the SCPNet in Fig. 13, we use the mean squared error (MSE) between the ground truth outputs of the SCP solver and predictions of SCPNet as the cost function and apply RMSprop  \cite{tieleman2012lecture} as the optimizer. For each iteration, the parameters for weight $w$ and bias $b$ are updated by the following equations:
	\begin{subequations}
		\begin{eqnarray}
		w & = & w - \alpha\frac{g_w}{\sqrt{s_{g_w} + \epsilon}}\\
		\label{e:w}
		b & = & b - \alpha\frac{g_b}{\sqrt{s_{g_b} + \epsilon}},
		\label{e:d}
		\end{eqnarray}
	\end{subequations}
	where $\alpha$ denotes learning rate, $g_w$ and $g_b$ is the gradient of training loss with respect to $w$ and  $b$, respectively. As $s_{g_w}$ may be a small number close to 0, $\epsilon$ is added to ensure numerical stability; generally it will be set to $1e^{-8}$. Additionally, $s_{g_w}$ and $s_{g_b}$ can be calculated by Eq. \ref{e:sdw} and Eq. \ref{e:sdb}:
	\begin{subequations}
		\begin{eqnarray}
		s_{g_w} & = & \beta_1 s_{g_w} + (1 - \beta_1) g_w^2\\
		\label{e:sdw}
		s_{g_b} & = & \beta_1 s_{g_b} + (1 - \beta_1) g_b^2,
		\label{e:sdb}
		\end{eqnarray}
	\end{subequations}
	where $\beta_1$ is the value of the momentum, which takes the past gradients into account to smooth out the steps of the gradient descent. In \cite{tieleman2012lecture}, it is recommended to set $\beta_1$ to 0.9, as it works very well for most applications, including our network (we tried different values of $\beta_1$ between 0.9 to 0.99, and the results are basically consistent).

	During the network training process, we first set the initial model weights following the XAVIER method \cite{glorot2010understanding}. 
	We then deploy the training process by randomly sampling one from all the 1080 training sequences. For each iteration in the sampled sequence, we backpropagate the total loss of 30 timesteps of data and we update the model weights as in (15) and (16). As a result, a sequence consisting of 1200-1650 timesteps leads to 40-55 iterations. For every 10 sequences (a sequence-batch), the learning rate is updated by decreasing linearly from $10^{-4}$ to $10^{-8}$. This procedure repeats until the loss is sufficiently small, as shown in Fig. \ref{fig:ErrorAngleAbx}. The final training loss is quite small, at the level of $10^{-4}$. We also notice that the training time is short, at 2-3 hours.    
	
	\subsection{Network testing}
	To evaluate the performance of the trained SCPNet, testing data are applied. Fig. \ref{fig:ErrorAngleAbx} shows the error distribution of the outputs, where the learning errors of pitch, roll, and yaw are mostly less than 2 degrees, and the errors of body X acceleration are less than 0.2 $\rm{m}/\rm{s}^2$. Additionally, Table \ref{tab:comutation_time_comparison} shows the comparison of the computation time of SCPNet and the SCP solver for the same 100,000 sets of testing data, including all the flight modes with and without disturbances. It shows that SCPNet has a significant advantage on the computation time over the SCP solver: the average and maximum computation times for the SCP solver is 1227.22 ms and 10482.91 ms, respectively; while the average and maximum computation time for SCPNet is 1.35 ms and 5.15 ms, respectively. For the same testing data as the inputs, the percentage for which the SCP solver failed to derive the solution within 0.02 s is as high as 99.82 \%, making it unsuitable for real flights. On the other hand, the same percentage for SCPNet is 0, thereby meeting the computation time requirements perfectly. 
	
	\begin{table}
		\centering
		\caption{ Computation time comparison of SCPNet and the SCP solver 
			(100,000 sets of testing data)}.
		\label{tab:comutation_time_comparison}
		\begin{tabular}{ccc}
			\toprule
			Metrics & SCPNet & SCP Solver  \\
			\midrule
			avg time (ms) & 1.35  & 1227.22  \\
			\midrule
			max time (ms) & 5.15   & 10482.91  \\
			\midrule
			time $>$ 0.02 s ($\%$)&0    & 99.82 \\
			\bottomrule
			
		\end{tabular}
		\vspace{-2mm}
	\end{table}
	
	Additionally, we test the network using a variety of trajectories such as hovering, takeoff, transition, forward flight, transition back and landing. Although data of transition back and landing are not included in the training data, testing results for these two parts are quite satisfactory. These results indicate the trained SCPNet has gained sufficient robustness to be generalized to un-seen scenarios. To streamline the manuscript, only the results for transition, forward flight, and the transition back part will be presented. 
	
	\begin{figure*}[t!]
		\centering
		\includegraphics[width=0.97\textwidth]{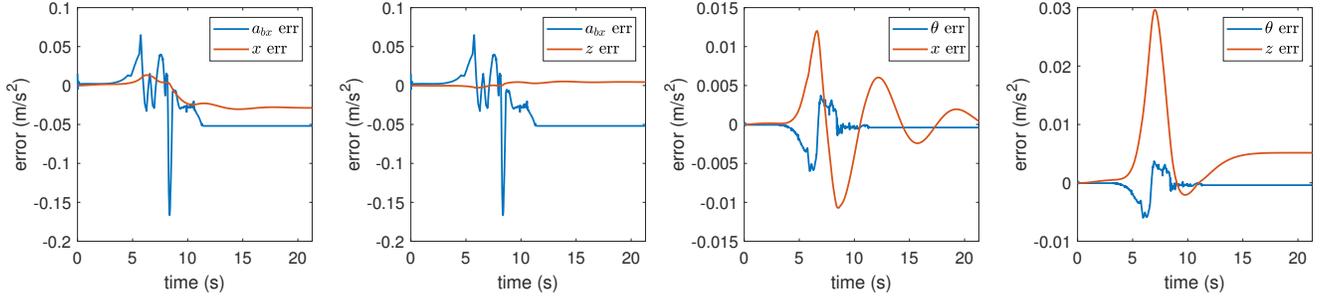}
		\caption{Tracking error caused by noise}
		\label{fig:noiseErr}
		\vspace{-0cm}
	\end{figure*}
	
	\begin{figure*}[t!]
		\centering
		\includegraphics[width=0.7\textwidth]{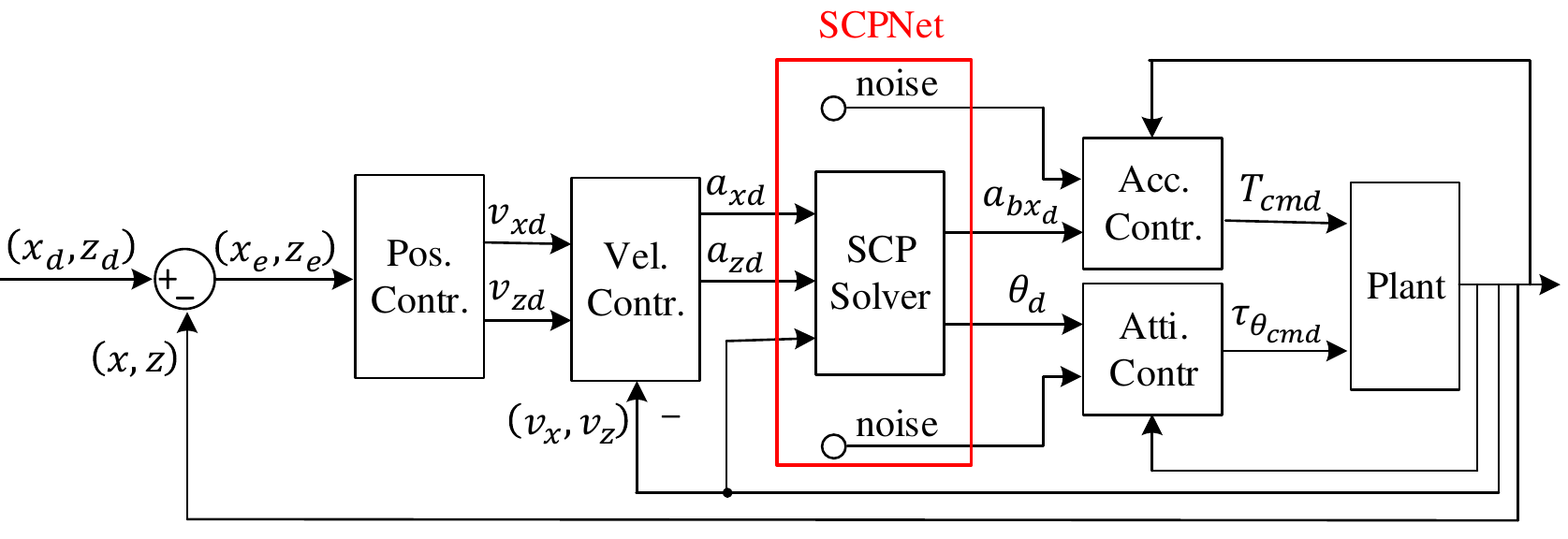}
		\caption{Closed-loop system in the longitudinal direction}
		\label{fig:sim_model_scpnet_xz}
		\vspace{-0cm}
	\end{figure*} 
	
	Fig. \ref{fig:scp_test} shows several testing results for SCPNet: the blue lines represent the outputs of the SCP solver, and the orange lines represent the outputs of SCPNet with the same inputs. The figures in the top row show the comparison of outputs between the SCP solver and SCPNet for transition and forward flight with no disturbance. We find the behavior of SCPNet is similar to that of the SCP solver; however, there always exist disturbances, such as wind, during real flights. Therefore, the figures in the middle row present testing results for the same trajectory with disturbance. The behavior of the SCPNet outputs is almost the same as that of the SCP solver. To better verify the ability of SCPNet, we test some un-seen scenarios during the training, such as transition back from forward flight, and the landing process. As the landing process is quite simple, here we only illustrate the transition back process, shown in the figures in the bottom row of Fig. \ref{fig:scp_test}.
	
	The learning accuracy is quite high, and even though no transition back data is involved in the training data set, the trained network has gained the main behavior of the SCP solver. Note that the behavior for SCPNet is smoother than for the SCP solver, especially for the timesteps between the transition back process and hovering. This is because the SCP solver only considers the current states and the last states for calculation of the outputs, while this neural network considers more historical states (the last states for five timesteps), thus improving the overall performance.

	\section{Software-in-the-loop Verification}
	
	Before the real implementation of this control method, we first analyze its stability and robustness, verify the effectiveness of this unified control method via a realistic simulator, and finally demonstrate its superiority by comparing to a conventional switching controller. 
	
	\subsection{Stability and robustness analysis }
	To analyze the stability and robustness of the designed neural network-based controller, we separate the longitudinal motion from the lateral one, which is proved to be decoupled in \cite{stevens2015aircraft}. In this subsection, we first show the detailed process of stability and robustness analysis in the longitudinal direction and then apply a similar analysis method to the lateral direction. 
	
	For the aircraft's longitudinal direction, the structure for the closed-loop system is shown in Fig. \ref{fig:sim_model_scpnet_xz}. Compared to Fig. \ref{fig:contrl_structure}, we only consider the 2D position $[x, z]$ and pitch angle $\theta$ instead of the 3D position $\bm{p} = [x, y, z]$ and the full attitude $\bm{R}$ described by $(\phi, \theta, \psi)$, and regard SCPNet as a combination of the SCP solver and noise. Because the noise does not affect the system's stability, the analysis of the stability of the closed-loop system with SCPNet is equivalent to that of the SCP solver. Additionally, the wind $\bm{w}$ and the desired velocity $\bm{v}_d$ are omitted in the inputs to the SCP solver. The wind is omitted and regarded as a disturbance, and the desired velocity is also omitted as it is only used in the hovering mode to provide a reference yaw direction for the aircraft. To analyze the stability and robustness, we linearize the aircraft dynamics and the designed controller (i.e., the SCPNet) as a whole, instead of deriving the linearized equations for each component (i.e., the SCPNet, and the aircraft dynamics) separately. The entire closed-loop system model is linearized in the frequency domain using the identification method proposed in \cite{zhou2018frequency}. Based on the identified model in the frequency domain, we use the Nyquist stability criterion in the frequency domain to analyze the stability \cite{doyle2013feedback}, and use the Bode plot diagram to determine the robust stability margins \cite{franklin2014feedback}. The details are shown as follows.
	
	\begin{figure}[t!]
		\centering
		\includegraphics[width=3in]{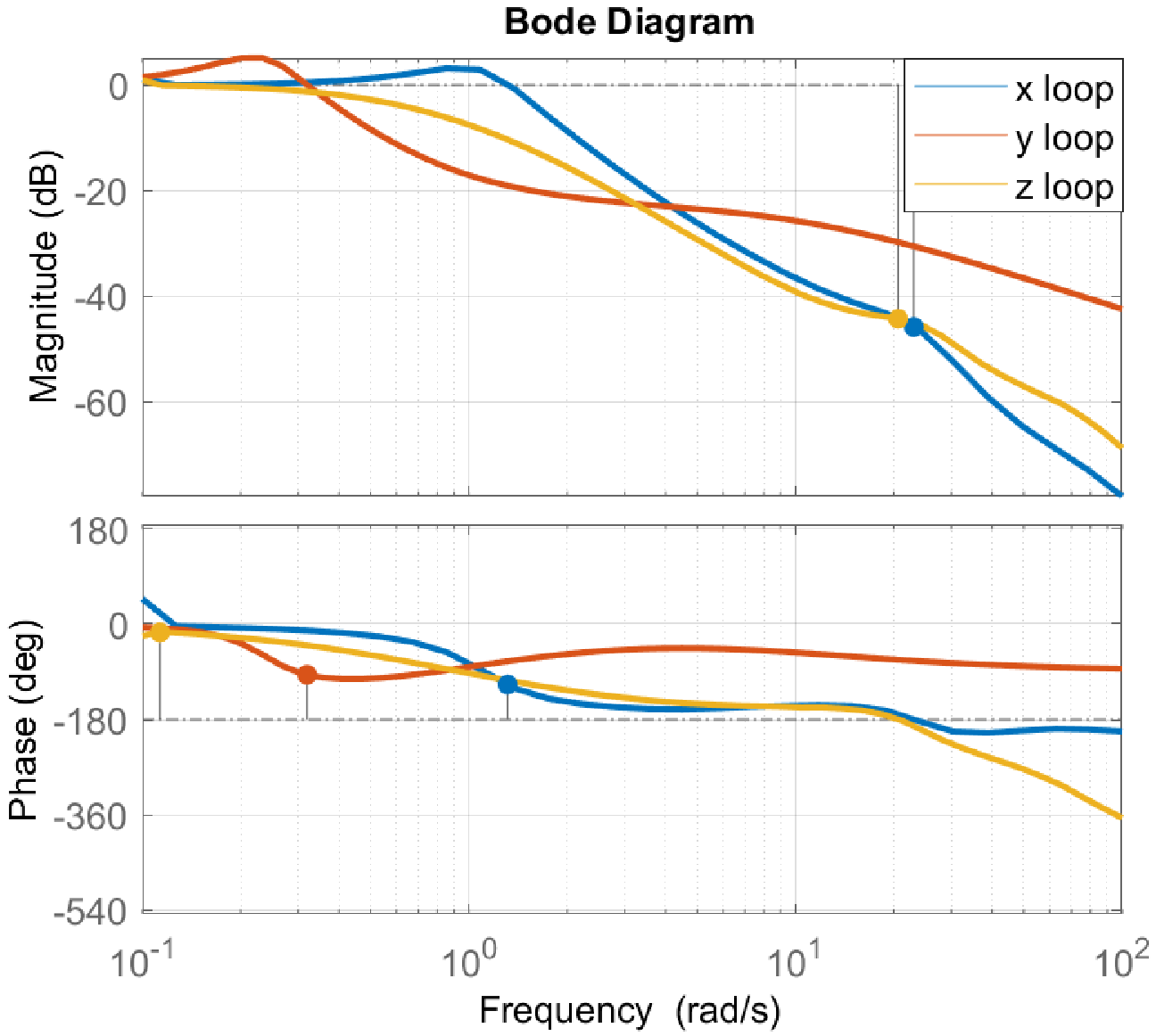}
		\caption{Bode diagram of the loop transfer functions in \\ the X, Y, and Z directions}
		\label{fig:bode_xyz}
		\vspace{-0cm}
	\end{figure}
	
	We linearize the open-loop system at different cruising speeds from 1 m/s to 10 m/s, which covers the states of takeoff, transition, level flight and landing. Because the wings of this aircraft have a small angular velocity aerodynamic damping coefficient, the dynamic results show consistency at different cruising speeds, so we select the frequency domain data at 10 m/s (optimal flight speed) as a representative for analysis. Fig. \ref{fig:bode_xyz} shows the identified loop transfer functions (the transfer function from $x_e$ and $z_e$ to $x$ and $z$, respectively, in Fig. \ref{fig:sim_model_scpnet_xz}), where we find that the gains when the phase crosses -180 deg are both smaller than 0 dB for the X and Z directions, indicating that the closed-loop systems along these two directions are stable by the Nyquist stability criterion. Moreover, the same figure also implies the robustness of the closed-loop system: the gain margins are 45.9 dB and 44.3 dB in X and Z directions, respectively, indicating the gain mismatches are up to 197.2 and 164.1 before being unstable, and the phase margins are 65.3 deg and 164 deg in X and Z directions, respectively, indicating the delay is up to 0.9 and 25.3 seconds before being unstable.
	
	Similarly, we analyze the stability and robustness of the aircraft's lateral direction. It is also proved stable by the Nyquist stability criterion, and  Fig. \ref{fig:bode_xyz} shows the Bode diagram of the loop transfer function in the Y direction, from which it can be read that the gain margin is $(-\infty, +\infty)$ dB, and the phase margin is 83.9 deg.

	Moreover, to analyze the influence of noise on the control accuracy, we calculate the output errors (i.e., noise in Fig. \ref{fig:sim_model_scpnet_xz}) between SCPNet and the SCP solver for the same input points of an entire trajectory, and use the output errors as inputs of the derived transfer functions to get the position tracking errors, shown in Fig. \ref{fig:noiseErr}. From the figure, we find the maximum error is 0.03 m in the Z direction, caused by the noise on  $\theta_d$. These errors are negligible in the actual flights and have a minor influence on the robustness margins. 
	
	\begin{figure}[t!]
		\centering
		\includegraphics[width=3.3in]{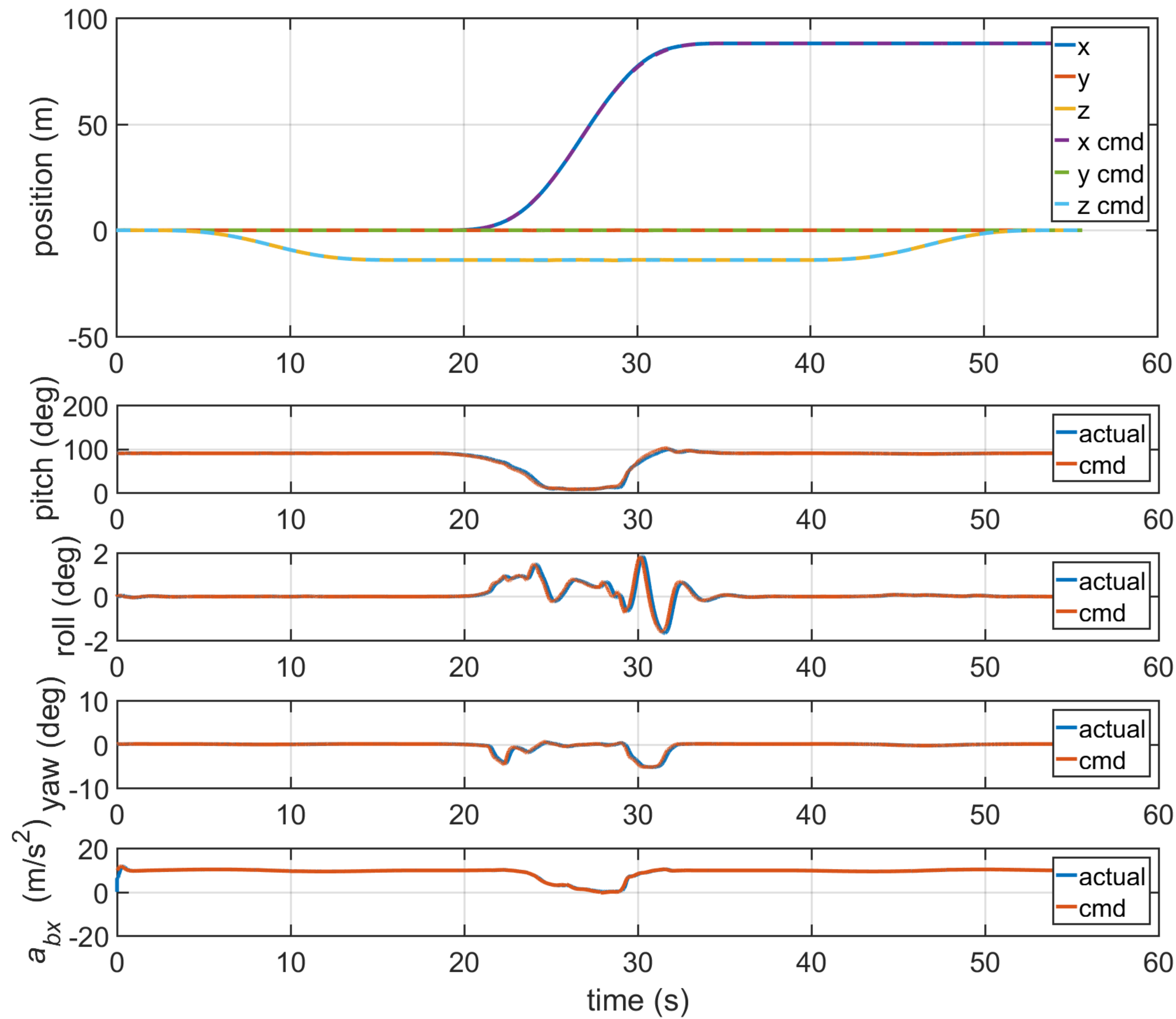}
		\caption{Position (top)  and attitude (bottom) response without disturbance for an entire trajectory (consisting of taking off, hovering, forward transition, level flight, backward transition and landing)}
		\label{fig:sim_pos}
		\vspace{-0cm}
	\end{figure}

	\begin{figure}[t!]
		\centering
		\includegraphics[width=3.5in]{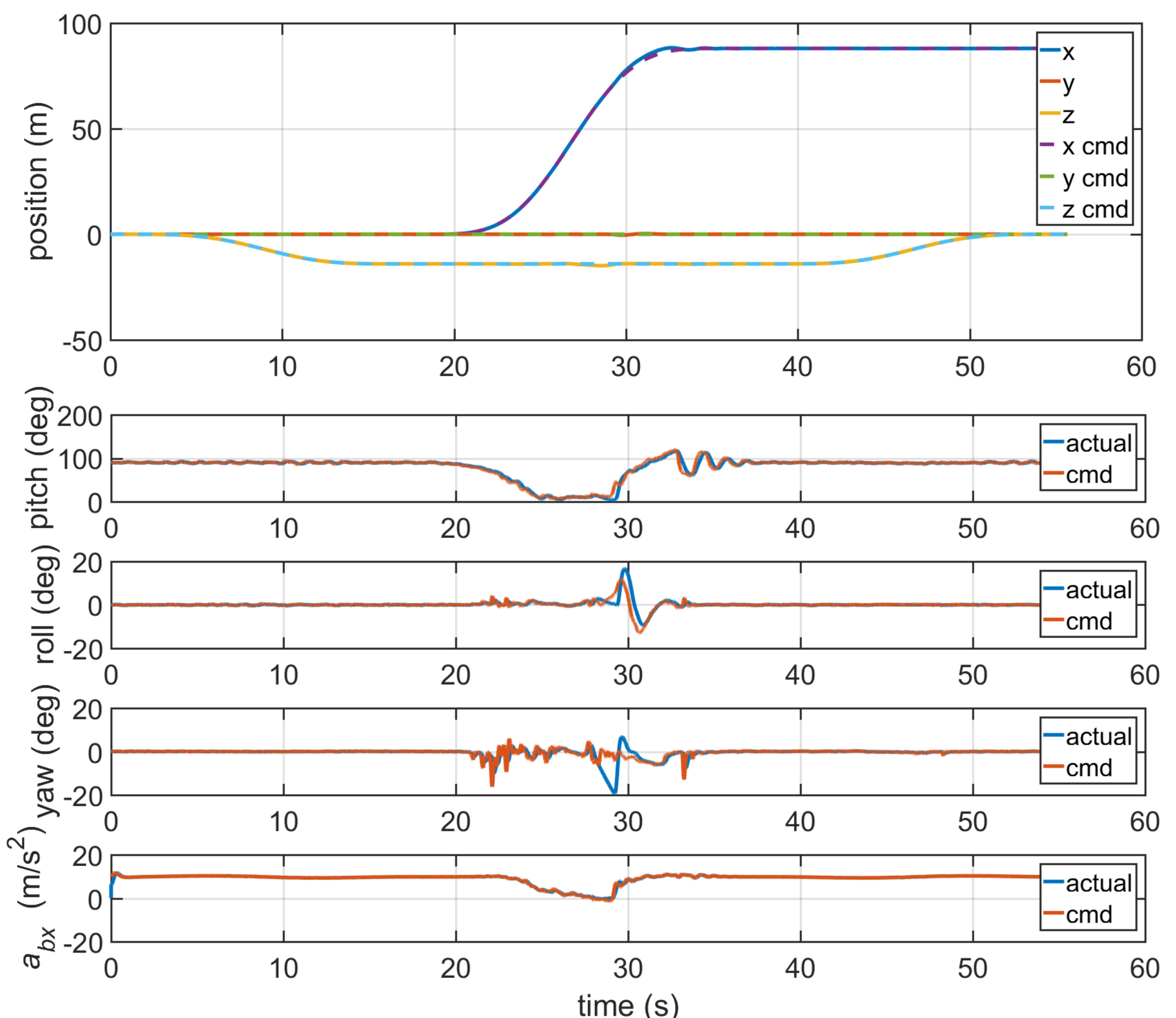}
		\caption{Position (top)  and attitude (bottom) response with disturbance for an entire trajectory}
		\label{fig:sim_dis_atti_pos}
		\vspace{-0cm}
	\end{figure}

	\subsection{Verification for typical trajectories}
	
	After the stability and robustness are proved to meet the requirements, we design several typical trajectories to analyze the performance both in no disturbance scenarios and in disturbance scenarios via a realistic simulator. The simulation structure is similar to that in Fig. \ref{fig:contrl_structure}, with the main difference being that there are no wind inputs to SCPNet because the accuracy of the onboard pitot tube for wind speed measurement is low. Therefore, we omit the wind inputs to SCPNet and regard the wind as a disturbance both in simulation and real flight experiments. 
	
	\subsubsection{Trajectories with no disturbance}
	
	For the desired trajectory $p_d$ including takeoff, transition, forward flight, transition back and landing process, in Fig. \ref{fig:sim_pos}, the top figure shows the position response, the solid line represents the actual position, and the dashed line represents the desired position. In this figure, the aircraft first takes off to the height of 14 m, transits to forward flight, transits back and stops at 88 m along the X direction, and finally, lands on the ground. The position is followed accurately as the curves representing the actual position and the desired position almost coincide with each other. 
	
	Additionally,  the bottom figure shows the commands generated by SCPNet and the command execution conditions. At first, the command for pitch, roll, and yaw is 90, 0, and 0 degrees, respectively, and the command for body X acceleration (represented as $a_{bx_d}$) is 9.8 $\rm{m}/\rm{s}^2$, which represents the hover condition. During the takeoff stage, there are minor changes to the commands. For the transition stage, the pitch angle smoothly changes from 90 to 5 degrees, and $a_{bx_d}$ command changes from 9.8 to almost 0 $\rm{m}/\rm{s}^2$. For the forward flight stage, the commands maintain stable values. Then comes the transition back and landing stage, which are the opposite form of transition and takeoff. During the entire process, the commands are reasonable and smooth with a vibration amplitude of less than 2 degrees, and the commands are executed well in the simulator.     
	
	\subsubsection{Trajectories with disturbance}
	
	In the previous part, we verified the control method with no disturbance, in this part, we will analyze the performance when disturbances are presented. The verification with disturbance is important before implementation as the wind usually exists and is regarded as a disturbance in real flights. 
	
	\begin{table}
		\centering
		\caption{ Position errors for typical scenarios \\
			(SCPNet in the simulation loop)}.
		\label{tab:err_analyze}
		\begin{tabular}{cccc}
			\toprule
			Trajectory & Metrics & Errors & RMSE \\
			\midrule
			
			entire trajectory\textsuperscript{$\ast $}
			&x (m)     & [-0.12, 0.45]  & 0.08  \\
			with no disturbance 
			&y (m)     & [-0.05, 0.03]  & 0.01  \\
			&z (m)     & [-0.16, 0.11]  & 0.03  \\
			\midrule
			entire trajectory\textsuperscript{$\ast $}
			&x (m)     & [-0.45, 1.78]  & 0.37 \\
			with disturbance
			&y (m)     & [-0.37, 0.32]  & 0.05 \\
			&z (m)     & [-0.86, 0.15]  & 0.12 \\
			\midrule
			hovering
			&x (m)     & [-1.17, 1.87]  & 0.64  \\
			with disturbance
			&y (m)     & [-0.98, 1.97]  & 0.62  \\
			&z (m)     & [-0.99, 1.86]  & 0.58  \\
			\bottomrule
			
		\end{tabular}
		{ \begin{flushleft}
				
				\hspace{1cm} \textsuperscript{$\ast $} includes taking off, hovering, forward transition,\\ \hspace{1.25cm} level flight, backward transition and landing
			\end{flushleft}
		}
		\vspace{-2mm}
	\end{table}

	For the same entire trajectory analyzed in the previous subsection, we add Gaussian white noise as the disturbance to the input of $\bm{v}$ in Fig. \ref{fig:contrl_structure}, and the amplitude of this disturbance is from -3 to 3, which is normal wind speed in the real experiment environment. Fig. \ref{fig:sim_dis_atti_pos} shows the position and attitude response for this scenario. Compared with Fig. \ref{fig:sim_pos}, we find there exists a much larger attitude vibration, but the scale is within a controllable rage (within 20 degrees). As a result, the position commands are followed well with the presented disturbances. Table \ref{tab:err_analyze} shows the error ranges (Errors) and root mean square error (RMSE) for each position in the x, y, and z directions. Here, three typical trajectories are analyzed. The position errors for the entire trajectory with disturbance are larger than those of the no disturbance trajectory, but the largest error is less than 1.78 m, with an RMSE of less than 0.37, which indicates the system has good performance with normal wind disturbance.    
	
	Moreover, we also verify the hovering scenario with disturbance. In this test, we use 100,000 timesteps of hovering trajectory points with a large random disturbance (amplitude from -6 to 6) added to $\bm{v}_i$. For this large disturbance, the position varies within a small range, less than 2 m, the details of which are shown in the last three columns of Table \ref{tab:err_analyze}. 
	
	\begin{figure}[t!]
		\centering
		\includegraphics[width=3in]{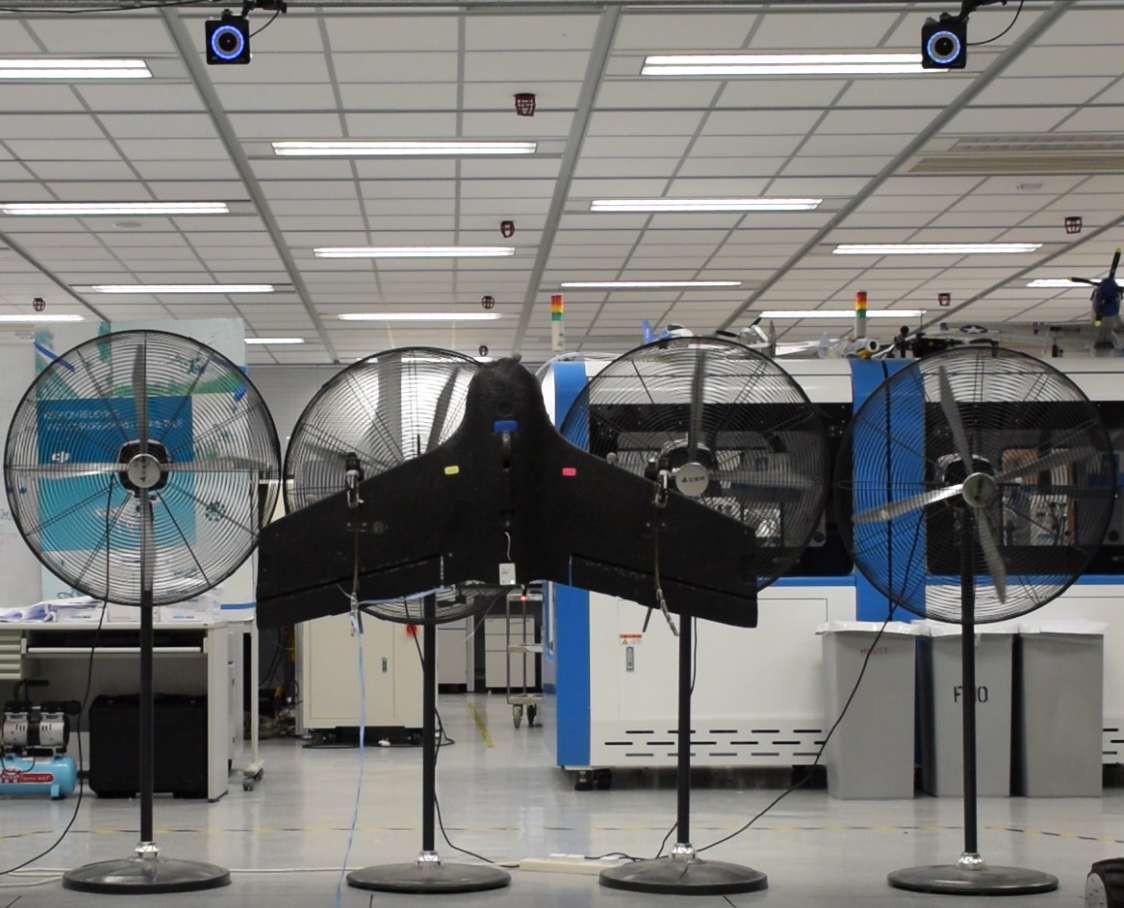}
		\caption{Environment of indoor experiments}
		\label{fig:indoor_exp}
		\vspace{-0cm}
	\end{figure}
	
	\subsection{Comparison with traditional control methods}
	As covered in Section II on related work, most of the existing methods employ a hierarchical control framework for tail-sitter VTOL UAVs, in which different controllers are used for different flight modes such as hovering, transition and level flight. Here we compare our design with one such control method taken from \cite{lyu2017hierarchical}. Note that the transition process (i.e., forward transition or back transition) therein is not satisfactorily smooth and the tracking accuracy is detrimentally affected as a consequence. For a forward transition from hovering mode to level flight mode, the aircraft executes the transition by sending a linearly decreasing pitch angle command to the altitude holding controller, while the roll and yaw angles are set to zero. Once the prescribed pitch angle and airspeed are both reached, the level flight mode will be triggered. Similarly, the aircraft in the level flight mode will execute the backward transition by calling the same altitude holding controller if a transition command is given. During the transition process, the aircraft only controls the altitude, leaving the position in the X and Y directions uncontrolled. Although the transition process only takes several seconds, it significantly affects the trajectory tracking accuracy, especially the lateral motion, when a disturbance (such as a wind gust) is present.  
	
	\begin{figure*}[t!]
		\centering
		\includegraphics[width=0.95\textwidth]{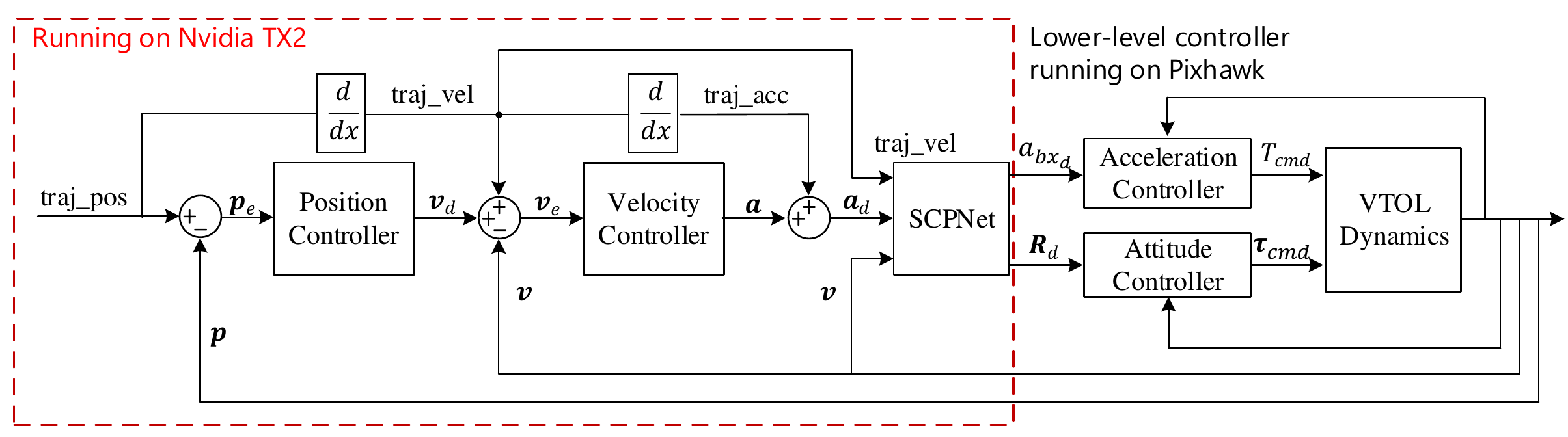}
		\caption{Experiment structure}
		\label{fig:expriment_structure}
		\vspace{-0cm}
	\end{figure*}
	
	To compare the performance between the hierarchical control framework and the unified control framework, here we focus on the forward transition process, for which the results are shown in Table \ref{tab:method_comparision}. In the absence of disturbance, their performances are similar. In practice, however, there always exists wind disturbance during flights, and we expect the aircraft to operate in normal weather when the wind speed is less than 5 m/s. Here, we generate random Gaussian noises with a variance of 5 $\rm{m}^2/\rm{s}^2$ as disturbances to each of the simulators. In the face of such disturbances, the unified framework has an obvious advantage on the position tracking performance, especially in the Y direction, where the maximum error is 1.4 m, while that of the hierarchical framework is 7.12 m.

	\begin{table}
		\newcommand{\tabincell}[2]{\begin{tabular}{@{}#1@{}}#2\end{tabular}}
		\centering
		\caption{ Transition process from hovering to level flight \\
			(i.e., pitch angle from 90 to 25 degrees)}.
		\label{tab:method_comparision}
		\begin{tabular}{cccc}
			\toprule
			Conditions & Metrics & \tabincell{c}{Unified \\ framework} & \tabincell{c}{Hierarchical \\framework} \\
			\midrule    
			no disturbance
			&x distance (m)     & 15.46  & 17.12  \\
			&y max error (m)     & 0.32  & 0.61  \\
			&z max error (m)     & 0.01  & 0.03  \\
			&time consumed (s)     & 3.32   & 3.21  \\
			\midrule
			with disturbance
			&x distance (m)    & 15.25  & 20.74 \\
			(random Gaussian noise
			&y max error (m)     & 1.4 & 7.12 \\
			variance: 5 $\rm{m}^2/\rm{s}^2$)
			&z max error (m)     & 0.57 & 1.39 \\
			&time consumed (s)     & 3.4   & 3.29  \\
			\bottomrule    
		\end{tabular}
		\vspace{-2mm}
	\end{table}
	
	\section{Experiment Results}
	Building upon the theoretical performance verification of the control method with SCPNet, we conduct extensive experiments to demonstrate the strength of our proposed framework. The control structure for the experiments is shown in Fig. \ref{fig:expriment_structure}. The position controller and SCPNet are run on an onboard computer, TX2, and the outputs of the SCPNet are packaged as MAVlink messages. The messages are sent to the pixhawk and received by the attitude controller and body X acceleration controller separately. The frequency of the position controller is 50 Hz and the attitude controller is 250 Hz.
	
	This section is divided into two parts: indoor experiments and outdoor experiments, and a video of these experiments are provided on the link "https://youtu.be/xm8CZagg6V8".

	\subsection{Indoor Experiments}
	For safety consideration, we first performed some indoor tests before conducting outdoor flight testing. Figure \ref{fig:indoor_exp} shows the environment of the indoor experiments. Here, a motion capture system is used to provide the position information of the aircraft, and four industrial fans are utilized to generate a non-uniform wind field imitating wind in the outdoor environment. A handheld anemometer measures the wind speed in the hovering position of the aircraft, and the value is between 3.5 - 4.1 m/s, which is common wind speed in Hong Kong.

\begin{figure}[t!]
		\centering
		\includegraphics[width=3.5in]{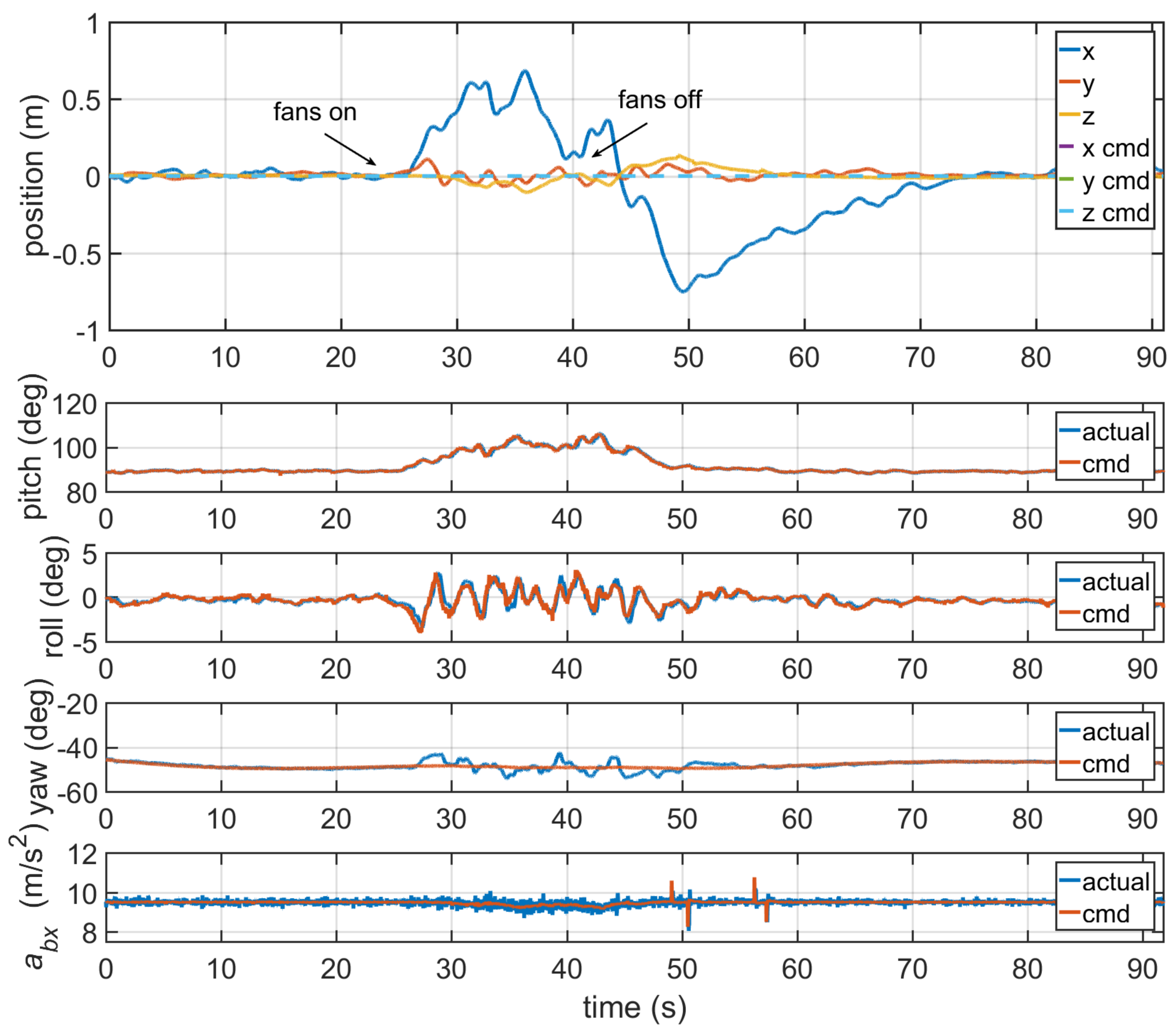}
		\caption{Position and attitude response for a hovering trajectory with disturbance}
		\label{fig:indoor_hover_pos}
		\vspace{-0cm}
	\end{figure}

	Figure \ref{fig:indoor_hover_pos} shows the position and attitude performance of these hovering experiments. From the top figure, we find the original position of the aircraft is [0, 0, 0], then the aircraft starts to drift when the fans are turned on. After the fans are turned off, the aircraft returns to the original point. The bottom figure shows the attitude and body X acceleration performance of this hovering experiment where there exist some sudden changes in attitude due to the change of the wind speed, but it can return to a stable attitude after the wind is stopped.
	
	Table \ref{tab:err_hover} shows the error range (Errors) and root mean square error (RMSE) for each of the position, attitude and body X acceleration: (1) the position errors are less than 0.75 m, and RMSE is less than 0.27 m; (2) the max attitude errors are less than 0.11 deg, and RMSE is less than 0.02 deg; and (3) the body X acceleration errors are less than 1.43 $\rm{m}/\rm{s}^2$, and RMSE is less than 0.13 $\rm{m}/\rm{s}^2$.   
	
	From this indoor experiment, we conclude that the position drift and attitude vibration are within a safe range (i.e., less than 1 m) when the wind speed is less than 4.1 m/s.

	\subsection{Outdoor Experiments}
	Finally, we conduct outdoor experiments when the wind speed is less than 4.1 m/s. This section will show the results of an entire trajectory including hovering, takeoff, forward transition, level flight, back transition, and landing. 
	
	The aircraft first takes off and climbs to a height of 14 m. The speed first changes from 0 $m/s$ to 2 $m/s$, then from  2 $m/s$ to 0 $m/s$ in the Z direction. After takeoff, the aircraft hovers for 2 seconds and then starts to transit from hovering to level flight. After it reaches the speed of 12 $m/s$, it maintains this speed for 4 seconds, then transits back to hovering, before at last landing on the ground.
	
	\begin{table}[t!]
		\centering
		\caption{ Position, attitude and body X acceleration errors for a hovering trajectory with disturbance.}
		\label{tab:err_hover}
		\begin{tabular}{ccc}
			\toprule
			Metrics & Errors & RMSE \\
			\midrule
			x (m)     & [-0.75, 0.68]  & 0.27  \\
			y (m)     & [-0.06, 0.11]  & 0.02  \\
			z (m)     & [-0.10, 0.13]  & 0.03  \\
			\midrule
			pitch (deg)     & [-0.05, 0.03]  & 0.01  \\
			roll (deg)     & [-0.03, 0.03]  & 0.01  \\
			yaw (deg)     & [-0.09, 0.11]  & 0.02  \\
			\midrule
			$a_{bx_d} (\rm{m}/\rm{s}^2)$     & [-1.39, 1.43]  & 0.13  \\
			\bottomrule    
		\end{tabular}
		\vspace{-2mm}
	\end{table}
	
\begin{figure}[t!]
		\centering
		\includegraphics[width=3.5in]{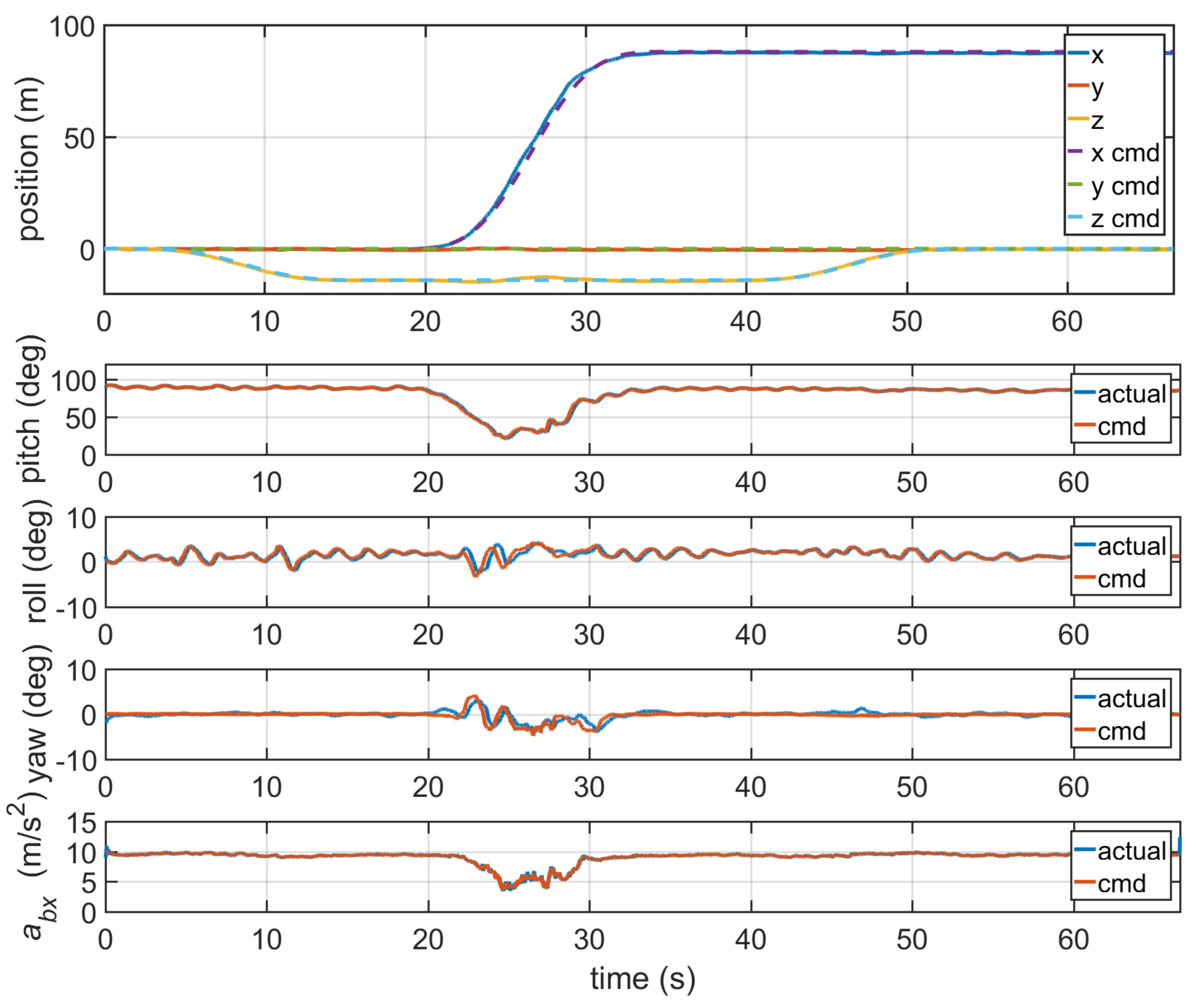}
		\caption{Position and attitude response for an entire trajectory}
		\label{fig:whole_pos}
		\vspace{-0cm}
	\end{figure}

	Fig. \ref{fig:whole_pos} shows the position and attitude response for the entire trajectory experiment. The performance is satisfactory with small drifting errors. From the figure, we find the position commands are executed well and the attitude performance is smooth. The initial yaw performance looks strange, and this is because the aircraft is not precisely facing the North direction, while the initial yaw command is set to 0 degrees in the controller. Therefore, it has a process to turn the yaw angle back to 0 degrees. 
	
	Overall, the performance is satisfactory with small and reasonable tracking errors due to the presence of wind disturbance.  The details are shown in Table \ref{tab:err_whole}: (1) the position errors are less than 3.28 m, and RMSE is less than 0.81 m; (2) the max attitude errors are less than 0.07 deg, and RMSE is less than 0.02 deg; (3) the body X acceleration errors are less than 3.02 $\rm{m}/\rm{s}^2$, and RMSE is less than 0.17 $\rm{m}/\rm{s}^2$. 
	
	\begin{table}[t!]
		\centering
		\caption{ Position, attitude and body X acceleration errors for an entire trajectory with disturbance.}
		\label{tab:err_whole}
		\begin{tabular}{ccc}
			\toprule
			Metrics & Errors & RMSE \\
			\midrule
			x (m)     & [-0.98, 3.28]  & 0.81  \\
			y (m)     & [-0.68, 0.20]  & 0.37  \\
			z (m)     & [-0.86, 1.40]  & 0.37  \\
			\midrule
			pitch (deg)     & [-0.16, 0.07]  & 0.02  \\
			roll (deg)     & [-0.06, 0.07]  & 0.01  \\
			yaw (deg)     & [-0.05, 0.07]  & 0.01  \\
			\midrule
			$a_{bx_d} (\rm{m}/\rm{s}^2)$     & [-1.03, 3.02]  & 0.17  \\
			\bottomrule    
		\end{tabular}
		\vspace{-2mm}
	\end{table}
	
	We find the trajectory tracking errors in real flight are larger than in the simulation, this is mainly caused by wind disturbance and the difference between the aircraft the simulator and the real aircraft.
	
	\section{Conclusions}
	In this paper, we proposed an RNN-based unified control framework for a tail-sitter VTOL UAV, which consists of an attitude controller, an acceleration controller, and an optimization-based position controller. Compared to other traditional control methods, this framework uniformly handles all the flight modes including taking off, hovering, forward/backward transition, level flight, and landing. Since the proposed non-linear optimization solver in the position controller may not be suitable for real-time implementation from a computational perspective, we apply an RNN to mimic the performance of a non-linear solver and employ it for real flights. In addition to efficient running speed and high approximation accuracy, the proposed RNN also eliminates the sudden changes in control actions computed by the non-linear solver, thereby mitigating the excitation of the flexible modes of the UAV. After analyzing the system stability and robustness, we conduct extensive simulations and experiments to show the effectiveness of this control method.
	
	Although this framework has already reached the maturity of real-world application, many directions for future research remain. The wind is regarded as a disturbance in this work, and the controllers are designed to work stably under certain disturbances. To improve the system robustness and trajectory tracking accuracy, we are interested in wind estimation methods to provide relatively accurate wind speed and disturbance observer design methodology to reduce the disturbance effects. Another research direction concerns the online generation of trajectories (including position, velocity, and acceleration) based on the aircraft dynamics, which also requires real-time obstacle avoidance algorithms working with appropriately assembled sensors.
	
	\section*{Acknowledgment}
	The authors would like to thank Dr. Sei Zhen Khong, Dr. Lei Tai, and Mr. Xiaoyu Cai for valuable discussions.
	

	\ifCLASSOPTIONcaptionsoff
	\newpage
	\fi
	
	\bibliographystyle{IEEEtran}
	\bibliography{paper}

\end{document}